\documentclass[10pt,twocolumn,letterpaper]{article}
\usepackage{iccv}
\usepackage{times}
\usepackage{epsfig}
\usepackage{graphicx}
\usepackage{amsmath}
\usepackage{amssymb}
 
\usepackage{multirow}
\usepackage{enumitem}
\usepackage{booktabs}
\usepackage[pagebackref=true,breaklinks=true,letterpaper=true,colorlinks,bookmarks=false]{hyperref}

 \iccvfinalcopy 

\ificcvfinal\fi

\begin{document}
\title{ Dual-Camera Super-Resolution with Aligned Attention Modules}
\author{Tengfei Wang$^1$\footnotemark[1] \quad

Jiaxin Xie$^1$\footnotemark[1]  \quad
 
Wenxiu Sun$^2$  \quad

Qiong Yan$^2$  \quad

Qifeng Chen$^1$

\\
\vspace{2mm}
{$^1$HKUST \quad $^2$SenseTime Research and Tetras.AI} 
}

\maketitle
\ificcvfinal\thispagestyle{empty}\fi
 \footnotetext[1]{equal contribution} 
\begin{abstract}
We present a novel approach to reference-based super-resolution (RefSR) with the focus on dual-camera super-resolution (DCSR), which utilizes reference images for high-quality and high-fidelity results. Our proposed method generalizes the standard patch-based feature matching with spatial alignment operations. We further explore the dual-camera super-resolution that is one promising application of RefSR, and build a dataset that consists of 146 image pairs from the main and telephoto cameras in a smartphone. To bridge the domain gaps between real-world images and the training images, we propose a self-supervised domain adaptation strategy for real-world images. Extensive experiments on our dataset and a public benchmark demonstrate clear improvement achieved by our method over state of the art in both quantitative evaluation and visual comparisons. Our code and data are available at \url{https://tengfei-wang.github.io/Dual-Camera-SR/index.html}.
\end{abstract}

\begin{figure*}[t]
    \centering
    \includegraphics[width=0.99\linewidth]{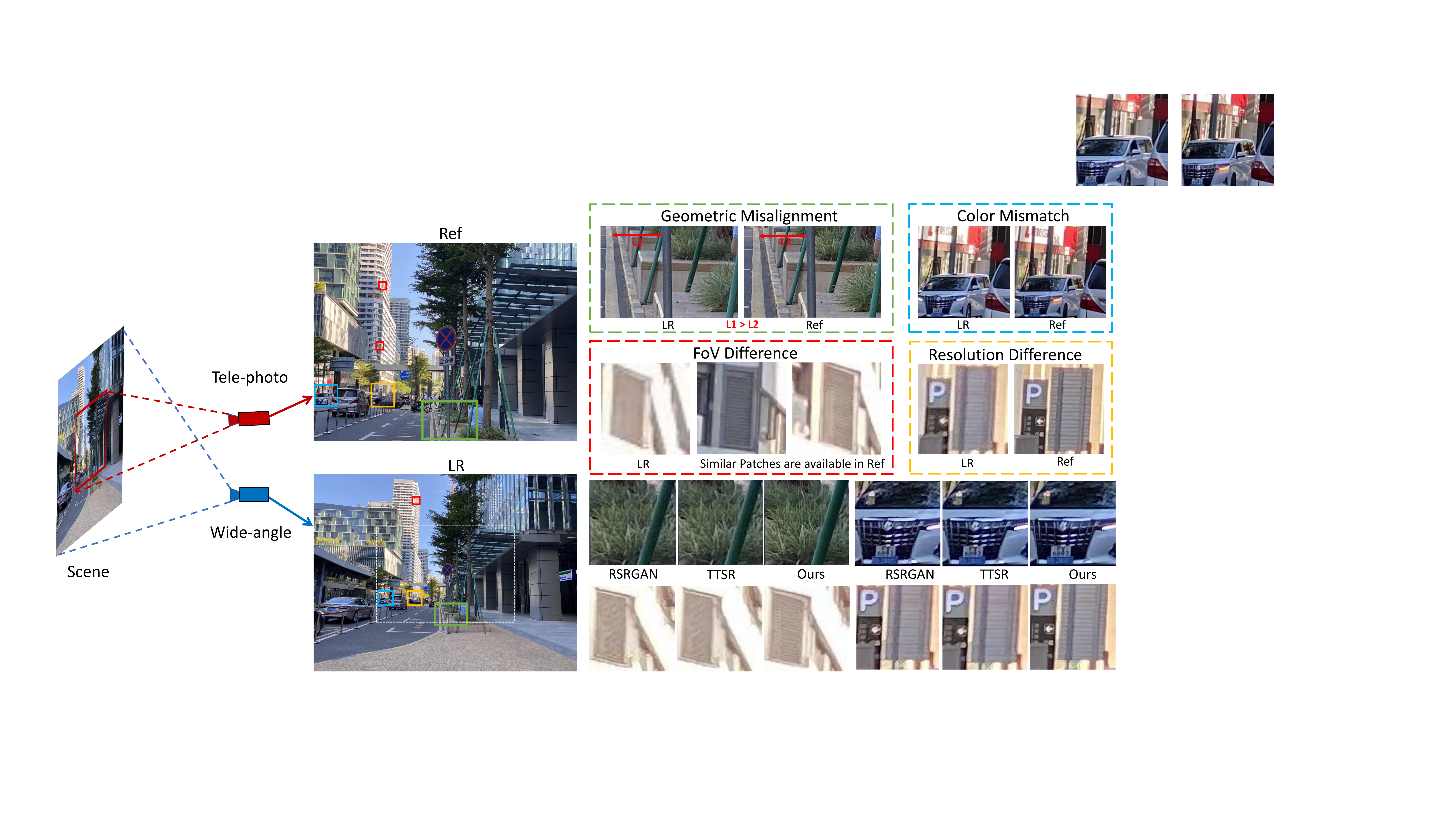}
    \vspace{-1mm}
    \caption{Demonstration of the smartphone dual-camera system. The telephoto and wide-angle images share  similar contents within the overlapped FoV (indicated by the white dotted line), while various misalignment exists. We take the telephoto image as the reference to super-resolve the wide-angle image for combining both large FoV and high-quality details. Compared with state-of-the-art SR approaches RSRGAN~\cite{zhang2019ranksrgan} and TTSR~\cite{yang2020learning}, our results are sharper and more realistic. Zoom-in for details.}
    \label{fig:teaser}
\end{figure*}

\section{Introduction}
Most smartphone manufacturers adopt an asymmetric-cameras system consisting of multiple fixed-focal lenses instead of a variable-focal one for optical zoom, due to limited assembly space. As shown in Fig.~\ref{fig:teaser}, the most common configuration has dual cameras with wide-angle (main camera) and telephoto lenses that have different field of views (FoV). The wide-angle and telephoto images often have spatial misalignment and color discrepancy due to viewpoint differences and different image signal processing (ISP) pipelines in the two lenses. As these two images capture the same scene with different focal lengths, can we use the telephoto image as a reference to enhance the resolution of the wide-angle image? To answer this question, we study reference-based super-resolution (RefSR) with the focus on dual-camera super-resolution (DCSR).

The key challenges of RefSR lie in (1) \textit{how to effectively  establish correspondences between low-resolution inputs (LR) and reference images (Ref)} (\textbf{feature warping}), and (2) \textit{how to integrate the reference information to improve the output image quality} (\textbf{feature fusion}). It has been widely observed that similar semantic patches and texture patterns tend to recur in the same or highly-correlated images with variable positions, orientations and sizes~\cite{Mei2020image,zontak2011internal}.  To search and utilize these correlated patterns  from reference images, previous learning-based approaches adopt either patch-wise matching (patch-match~\cite{zheng2017learning, zhang2019image}, patch-based attention~\cite{xiefeature, yang2020learning}) or pixel-wise alignment (optical-flow~\cite{zheng2018crossnet}, offsets~\cite{shim2020robust}), with different pros and cons. The pixel-wise alignment is able to handle non-rigid transformation, but  usually less stable and prone to generate  distorted structures due to the difficulty of reliable flow or offsets estimations~\cite{chan2020understanding}, especially for largely misaligned reference images.  Patch-wise matching  can achieve compelling warping performance since it evaluates similarity scores between LR and Ref patches in an explicit fashion. However, the vanilla patch-level matching lacks robustness to spatial misalignment, e.g. scaled or rotated patches. As shown in Fig.~\ref{fig:teaser}, even though highly-similar patches are available in the reference image, previous approaches  are insufficient to make use of these  cues, and tend to average the misaligned Ref  and LR patches to produce  blurry images. 

Another limitation of previous RefSR approaches is that they are difficult to be directly applied to  high-resolution images captured by smartphones. The reference images in  RefSR datasets~\cite{zhang2019image} are typically smaller than $512\times512$. Most   methods thus globally searches over the entire reference image for super-resolution cues. Nevertheless,  the memory  consumption of a global searching strategy would be intractable for the high-resolution cases (e.g. 4K).  The domain gaps between real-world images and  training images can also degrade the zoom performance~\cite{gu2019blind,zhang2019zoom,cai2019toward}.

To tackle these issues, we propose a deep RefSR method  with the focus on  dual-camera super-resolution. First, we generalize the vanilla patch-based attention to an aligned attention module, which  searches for related patches based on explicit matching, while implicitly learning  inter-patch transformations to alleviate spatial misalignment. Second, to prevent the reference patches from idling and contributing less to the super-resolution results, we  impose a  fidelity loss on the reference images. To advance our method to real-world images, we also propose a self-supervised adaptation strategy.  The main contributions of our paper can be summarized as:

\begin{itemize}[noitemsep,topsep=0pt]
  \item We are the first to explore  the real-world dual-camera super-resolution (wide-angle and telephoto cameras). We propose a  self-supervised domain adaptation scheme to bridge domain gaps between real-world images and downsampled images.
  
  \item We propose the aligned attention module and adaptive fusion module to improve the RefSR architecture. Our method outperforms state-of-the-art approaches  qualitatively and quantitatively.
  
  \item We  argue the importance of imposing an explicit fidelity loss  on reference images and performing explicit high-frequency fusion in the image space to the super-resolution quality.
\end{itemize}

\section{Related Work}
\subsection{Single Image Super Resolution}
 SISR~\cite{glasner2009super} has been actively explored  in recent years. After SRCNN~\cite{dong2015image},   MDSR~\cite{Lim_2017_CVPR_Workshops} introduced  residual blocks  to super-resolution area. RCAN~\cite{zhang2018rcan} further improved residual blocks by  channel attention.  To improve the perceptual quality, Johnson et al.~\cite{johnson2016perceptual} proposed the perceptual loss to minimize the feature distance. SRGAN~\cite{ledig2017photo} adopted generative adversarial networks~\cite{goodfellow2014generative} for more realistic textures. ESRGAN~\cite{wang2018esrgan} enhanced SRGAN with Residual-in-Residual Dense Block. RankSRGAN~\cite{zhang2019ranksrgan} combined SRGAN with a well-trained ranker that gives   ranking scores.  CSNLN~\cite{Mei2020image} proposed cross-scale non-local attention to find self-similarity for high-quality reconstruction.

\begin{figure*}[t]
    \centering
    \includegraphics[width=0.95\linewidth]{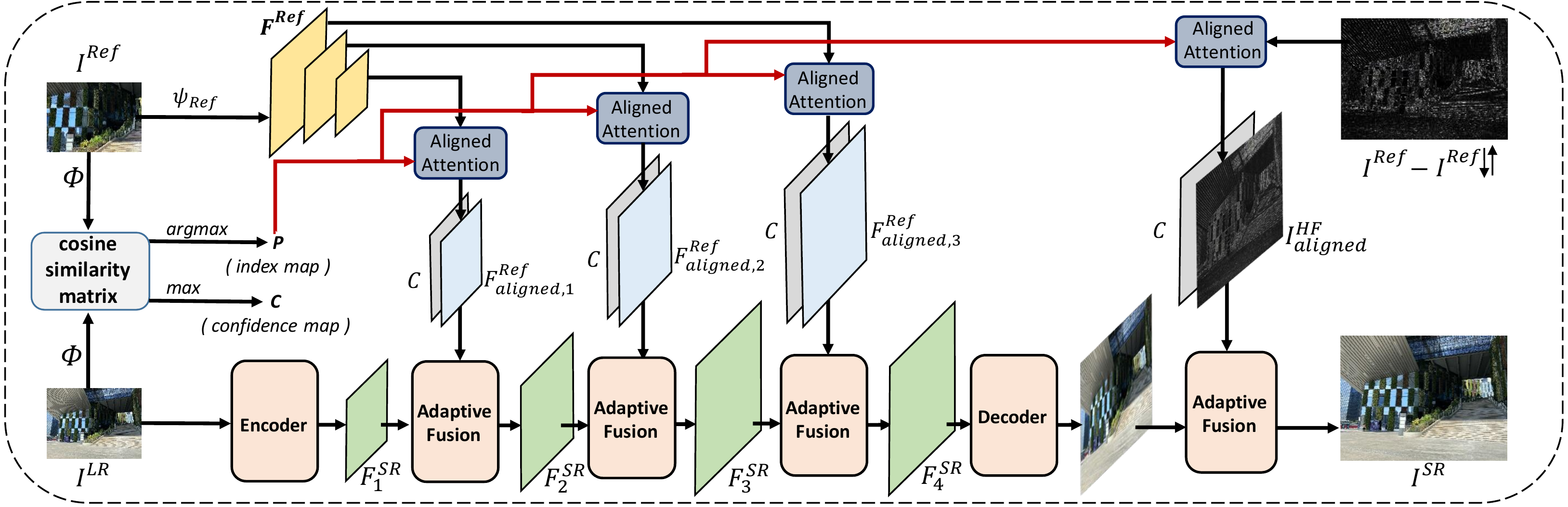}
    \vspace{1mm}
    \caption{Overview of our approach. We first match the nearest $I^{Ref}$ patch  for each $I^{LR}$ patch under cosine distance in feature space $\Phi$. The reference features $F^{Ref}$ at different scales are then warped by the aligned attention and adaptively fused with SR feature $F^{SR}$ according to this matching. After the final fusion with high-frequency $I^{HF}_{aligned}$, the network yields a high-fidelity output $I^{SR}$.}
    \label{fig:architecture}
\end{figure*}

\subsection{Reference-based Super Resolution}
RefSR alleviates the ill-posed nature of SISR by providing high-resolution reference images. Previous  learning-based approaches adopt either patch-wise matching  or pixel-wise alignment for feature warping. Pixel-wise alignment methods usually build a dense corresponding map, and warp the reference feature maps pixel by pixel. Zheng et al.~\cite{zheng2018crossnet} proposed to estimate optical flow between input and reference images  to warp feature maps at different scales.  However, it remains a challenging problem for reliable flow estimation in largely misaligned regions. Shim el al.~\cite{shim2020robust} proposed to implicitly estimate the offsets  with deformable convolution\cite{dai2017deformable} instead of optical flow.  The offsets warping is faster and more flexible than the  flow counterpart, while it is typically less stable.

Patch-wise matching  searches for  related patches by calculating similarity scores explicitly, which is thus more stable with better interpretability. Zhang et al.~\cite{zhang2019image} adopted Patch Match~\cite{barnes2009patchmatch} to warp features extracted by a  pretrained VGG network~\cite{simonyan2014very}. With a fixed VGG network as feature extractors, their method does not train the extractor jointly with the reconstruction net.  Yang et al.~\cite{yang2020learning} and Xie et al.~\cite{xiefeature} further proposed to adopt a learnable extractor and replace Patch Match with a patch-based attention, which allows an end-to-end learning pipeline. These patch-level warping methods can find semantic-similar patches, but are non-robust to inter-patch misalignment (e.g. scaled and rotated patches), which typically leads to blurry results. To address this issue, we proposed the aligned attention module that robustly warps spatially misaligned patches by estimating patch-wise alignment parameters.

\subsection{ Dual Camera Super Resolution}
The dual-camera super-resolution aims at super-resolving the wide-angle image with the telephoto image as a reference, which  combines both large FoV of short-focal camera and high resolution of long-focal camera. Most related works adopt traditional global correctness and registration techniques. Park et al.~\cite{park2016brightness} and Liu et al.~\cite{liu2014photometric} assumed that there is no disparity between wide-angle and telephoto pairs, and only correct  brightness and color globally. Some prior work considered the geometric misalignment between inputs and references. They simulated tele-image by center-cropping HR and  performing random affine transformation, and formulate this task to image registration. Yu et al.~\cite{yu2018continuous} applied RANSAC algorithm~\cite{fischler1981random} on SURF~\cite{bay2006surf} features to conduct global registration. Manne et al.~\cite{manne2019asymmetric} applied FLANN algorithm~\cite{muja2009fast} on ORB  features~\cite{rublee2011orb} for geometric registration. Nevertheless, there are huge domain gaps between the real-world telephotos and the simulated ones~\cite{gu2019blind, cai2019toward}, and previous approaches usually show significant performance drop in the  practical configuration. Instead of global image registration, We formulate DCSR as a  setting of  RefSR, and propose an end-to-end pipeline and training strategy. To the best knowledge of ours, we are the first learning-based method for real-world dual-camera super-resolution.
 
\begin{figure*}
    \centering
     \includegraphics[width=0.92\linewidth]{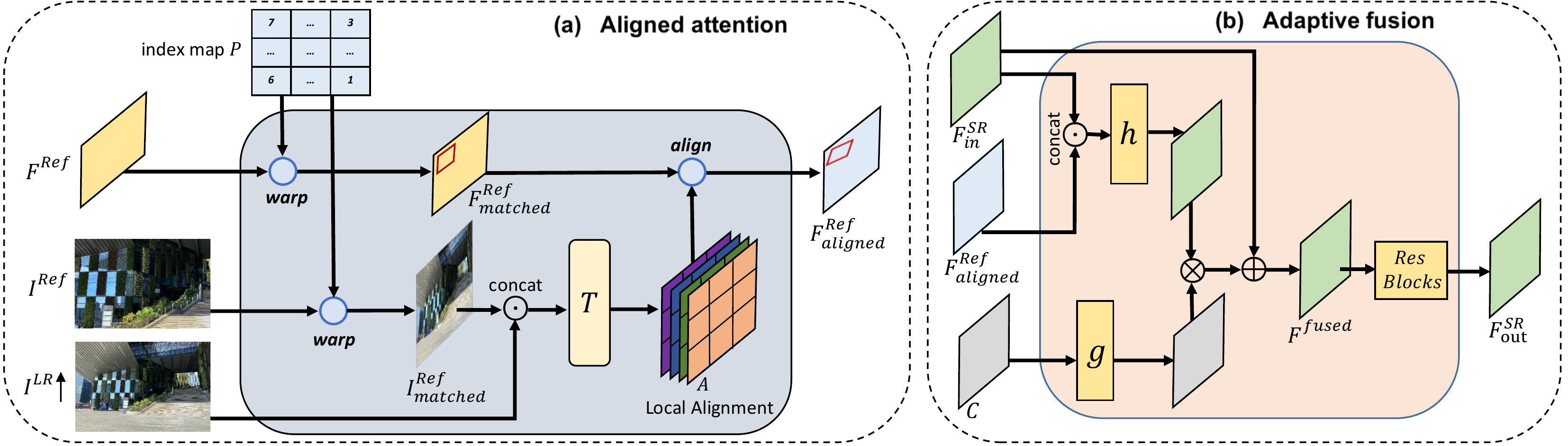}
     \vspace{1mm}
    \caption{ Illustration of the aligned attention module and adaptive fusion module. (a). The aligned attention applies index map $P$ to coarsely warp $F^{ref}$  and then fine-aligns  the patch   with the learned local transformation. (b). The adaptive fusion applies an additional  convolution layer  $g$  to aggregate  neighbor information in the confidence map $C$. }
  \label{fig:attention}
\end{figure*}

\section{Method}
Given $I^{LR}$ and $I^{Ref}$, we aim at generating a high-resolution image $I^{SR}$ that possesses high-quality details conditioned on $I^{Ref}$. As shown in Fig. \ref{fig:architecture}, our end-to-end  pipeline consists of two parts:  feature warping with aligned attention modules (Section~\ref{aa}), and feature fusion with adaptive fusion modules (Section~\ref{af}). To utilize both high-level and low-level information provided by $I^{Ref}$, following previous works~\cite{zhang2019image,yang2020learning}, we  extract reference features $F^{Ref}$ at different scale levels via encoder $\psi_{Ref}$. At each scale, we perform the aligned attention on  $F^{Ref}$ to warp it to match the LR for later fusion. This module can robustly match correlated patches and further align these patches to alleviate the differences in orientations and scales. After that, the aligned features $F^{Ref}_{aligned}$ as well as the high-frequency residual $I^{HF}_{aligned}$ are sequentially integrated with LR information guided by the matching confidence.

\subsection{Feature Warping with Aligned Attention }
\label{aa}
 Our method stems from the observation that similar patches tend to recur across correlated images  with  different scales and orientations~\cite{barnes2010generalized}.  The aligned attention aims at searching for these related reference patches and warp them to align with the LR counterparts. Following~\cite{zhang2019image}, we first perform a patch-wise matching~\cite{chen2016fast} to coarsely warp the reference, which is briefly reviewed below. $I^{LR}{\uparrow}$ and $ I^{Ref}$ are first embedded into feature maps  via a shared encoder $\phi(\cdot)$, and densely (stride=1) divided to $3 \times 3$  patches, where  ${\uparrow}$ denotes bicubic upsampling. We then calculate the cosine distance $S_{i,j}$ between each pairs of LR-patch i and Ref-patch j. For each LR-patch, we want to select the most relevant Ref-patch  for later feature fusion. The index map $P$ and confidence map $C$ of the matching are obtained as:

\begin{align}
P_i &=\underset{j} {\operatorname{arg} \operatorname{max}} S_{i,j}, 
&
C_i = \underset{j} { \operatorname{max}} S_{i,j}.
\end{align}  
 
The index map indicates the most relevant Ref-patch-$P_i$ for each LR-patch-$i$, and the confidence map gives the matching confidence $C_i$ for this match. The reference patches  can be warped now according to the index map, to obtain the coarsely matched images $I^{Ref}_{matched}$ and features $ F^{Ref}_{matched}$.

 Such an easy matching scheme~\cite{zhang2019image,chen2016fast,yang2020learning} performs stably on searching for similar patches. However, as shown in Fig.~\ref{fig:teaser}, even though highly-similar patches are available in $I^{Ref}$, there usually exists misalignment in orientation and scales. So far the coarsely-warped reference is not robust to rotation and scaling, which may yield blurry outputs by  averaging  unaligned Ref and LR.  Inspired by~\cite{barnes2010generalized, jaderberg2015spatial,dai2017deformable}, we propose to  estimate  patch-wise spatial transformation $A$ to further align  all matched patches in  $ F^{Ref}_{matched}$:

\begin{equation}
 A =  T  ( concat( I^{LR}{\uparrow} , I^{Ref}_{matched})).
\end{equation}
Instead of predicting a global transformation like~\cite{jaderberg2015spatial} for the whole $ F^{Ref}_{matched}$, the local spatial transformer network $T$ is designed to estimate patch-wise alignment parameters for all patches.  Each patch of $ F^{Ref}_{matched}$ is then aligned independently with the estimated affine matrix to  get the fine-aligned reference features $F^{Ref}_{aligned}$. $F^{Ref}_{aligned}$ will be used to facilitate the $I^{SR}$ generation by feature fusion.

\subsection{Adaptive Feature Fusion}
\label{af}
 A direct feature fusion (e.g. concatenation, summation) fails to consider the quality of the matches, which can inevitably bring irrelevant or noisy information. Prior work~\cite{yang2020learning} thus adopted the  confidence map  as a guidance for feature fusion. But in the original confidence map, $ C_i$  is calculated independently for each patch $i$, which means it  reflects the local matching confidence of every single
    patch, and the transition among neighbor patches is not necessarily smooth. To solve this issue, we  embed the confidence map with an extra convolution net $g$. It is a simple and effective way to aggregate neighbor confidences for more consistent and higher-quality results. The feature fusion process can be represented as:
\begin{equation}
    F^{fused} = g(C)\cdot h (F^{SR},F^{Ref}_{aligned}  ) + F^{SR},
\end{equation} where $g(\cdot)$ and  $h(\cdot)$ are learnable convolution layers. 

Another issue is that  the images reconstructed from the fused features tend to lose the high-frequency details. Inspired by recent work~\cite{yi2020contextual}, which generates high-frequency details by adding back image residuals with attention maps, we also conduct adaptive fusion in the image space. The aligned high-frequency (HF) residuals can be represented by $I^{HF}_{aligned} =   (I^{Ref} -  I^{Ref}{\downarrow}{\uparrow})_{aligned}$. However, different from inpainting task~\cite{yi2020contextual} where the HF details have no constraints in the missing regions, in super resolution the details need to be consistent with the original LR contents. To avoid introducing high-frequency noise, we also use a learnable function $g_r$ on the final fusion:
\begin{align}
\label{eq:image_sapce}
  I^{SR} &= g_r( C)\cdot I^{HF}_{aligned} +  decoder (F^{SR}).
\end{align}

\subsection{Loss Function}
We generate the output image conditioned on $I^{Ref}$, and expect $I^{SR}$  to approximate the ground-truth $I^{HR}$. Due to  misalignment between $I^{HR}$  and $I^{Ref}$, we found that using $I^{HR}$ as strict labels for the supervised learning leads to unsatisfactory details. We thus adopt the reconstruction term proposed in~\cite{ mechrez2018maintaining,mechrez2018contextual}, which calculates losses in  low-frequency and high-frequency bands separately :
\begin{align}
&\mathcal{L}_{rec}=\left\|I^{S R}_{blur}-I^{H R}_{blur}\right\|  + \sum_{i}  \mathrm{\delta}_{i }(I^{S R}, I^{H R}) ,
\end{align}
where $I_{blur}$ is filtered by $3\times3$ Gaussian kernels with $\sigma=0.5$. $\mathrm{\delta}_{i }(X, Y)= \min _{j}\mathbb{D}_{x_{i}, y_{j}}$ is distance between SR pixel  $x_i$ and its most similar HR pixel $y_i$  under certain distance $\mathbb{D}$~\cite{mechrez2018contextual,zhang2019zoom}. The first term  softly makes  $I^{SR}$ keep the same content as $I^{HR}$ in low-frequency domain. The second term flexibly enforces the statistics of $I^{SR}$  similar to   $I^{HR}$.

We find only using aforementioned losses yields blurry results, as the losses do not involve constrains on $I^{ Ref}$. Intuitively, the fusion modules in Fig~\ref{fig:attention} (b) can  easily ignore the reference information, and degrade to an identity mapping. In this case, $I^{ Ref}$ contributes less to $I^{ SR}$ generation. To avoid the `idleness' of $I^{ Ref}$, we introduce a fidelity term modified from~\cite{mechrez2018contextual}, where $\delta_{i}$ is the distance between   $I^{ SR}$ and  nearest-neighbor pixels in $I^{ Ref}$ under distance $\mathbb{D}$:
\begin{equation}
\mathcal{L}_{fid}= \frac{\sum_{i} \mathrm{\delta}_{i}(I^{S R}, I^{Ref}) \cdot c_i}{\sum_{i} c_i}. 
\end{equation}
Pixels with higher matching confidence $c_i$ are given larger weights for optimization, since these pixels can find highly-related cues in $I^{ Ref}$. This fidelity loss can adaptively maximize the similarity between $I^{ SR}$ and $I^{ Ref}$.
The overall loss  is the weighted sum of $\mathcal{L}_{rec}$ and $\mathcal{L}_{fid}$.

\subsection{Self-supervised Real-image Adaptation (SRA)}
\label{da}
For the real-world DCSR in Fig.~\ref{fig:teaser}, we take the  wide-angle image $I^{ wide}$ and telephoto image $I^{ tele}$  as $I^{ LR}$ and $I^{ Ref}$, respectively. We aim at super-resolving $I^{ wide}$  to produce $I^{ SR}$, but the ground-truth $I^{ HR}$ is unavailable to calculate the aforementioned losses. A typical training setting is to downscale the original  $I^{ wide}$ and $I^{ tele}$ by half to simulate the training inputs, and regard original $I^{ wide}$ as $I^{ HR}$ for supervised learning. However, we found that models trained on downsampled images show significant performance drop on the real  images (original $I^{ wide}$ and $I^{ tele}$) due to the domain gap between downsampled  and real-world images. To bridge this gap, inspired by recent works~\cite{ZSSR,ulyanov2018deep,Wang2021image}, we propose a  self-supervised real-image adaptation strategy (SRA) to fine-tune the trained model $M$  with real-world inputs without ground-truth. Specifically, we directly take the original $I^{ wide}$ and $I^{ tele}$ from the training set as  $I^{ LR}$ and $I^{ Ref}$, and  the training loss is defined as:
\begin{align}
&\mathcal{L} =\left\|I^{S R}\downarrow -I^{wide}\right\|  +\lambda \mathcal{L}_{fid} (I^{ SR}, I^{ tele} )
\end{align}
The first term  enforces $I^{ SR}$ to preserve the content of $I^{ wide}$, while the second term is to transfer $I^{ tele}$ details. After this training stage, the model $M' = \min_{M}\mathcal{L}$ generalizes well to the real-world inputs.

\section{Experiments}
\subsection{Datasets }
\textbf{CUFED5}~\cite{zhang2019image} It contains  11,871 training pairs and 126 test images. Each test image is accompanied with four reference images ranked by the similarity levels. The resolutions of HR and Ref are about 300$\times$500.

\textbf{CameraFusion } We construct a new dataset for dual-camera super-resolution, which contains 146 pairs of 4k wide-angle and  telephoto images in diverse outdoor and indoor scenes. As show in Fig.~\ref{fig:teaser}, they share the same scene but differ in  ISP and view-point.  A compelling RefSR approach is expected to show significant advantages over SISR methods  in the overlapped FoV area, while achieving comparable or better performance otherwise.

\begin{table}[t]
\small
\begin{minipage}[b]{37mm}
\begin{tabular}{@{}l@{\hspace{2mm}}c@{\hspace{2mm}}c@{}}
\hline
\textbf{SISR}  & PSNR & SSIM  \\
\hline
SRCNN~\cite{dong2015image}&25.33&0.745 \\
MDSR~\cite{Lim_2017_CVPR_Workshops}&25.93&0.777\\
RDN~\cite{zhang2018residual}&25.95&0.769\\
RCAN~\cite{zhang2018rcan}&26.06&0.769\\
LapSRN~\cite{lai2017deep}&24.92&0.730\\
SRGAN~\cite{ledig2017photo}&24.40&0.702\\
ENet~\cite{sajjadi2017enhancenet}&24.24&0.695\\
ESRGAN~\cite{wang2018esrgan}&21.90&0.633\\
RSRGAN~\cite{zhang2019ranksrgan}&22.31&0.635\\
CSNLN~\cite{Mei2020image}&24.73&0.743\\
\hline
\end{tabular}
\end{minipage}
\hspace{5mm}
\begin{minipage}[b]{37mm}
\begin{tabular}{@{}l@{\hspace{2mm}}c@{\hspace{2mm}}c@{}}
\hline
\textbf{RefSR}  & PSNR & SSIM  \\
\hline
Landmark~\cite{yue2013landmark}&24.91&0.718\\
CrossNet~\cite{zheng2018crossnet}&25.48&0.764\\
SRNTT~\cite{zhang2019image}&25.61&0.764\\
SRNTT-\textit{$\ell_2$}~\cite{zhang2019image}&26.24&0.784\\
SSEN\cite{shim2020robust}&26.78&0.791\\
FRM\cite{xiefeature}&24.24&0.724\\
TTSR~\cite{yang2020learning} &25.53&0.765\\
TTSR-\textit{$\ell_1$}~\cite{yang2020learning} &27.09&0.804\\
Ours &25.39 &0.733 \\
 \textbf{Ours-\textit{$\ell_1$}}&\textbf{27.30}  & \textbf{0.807}\\
\hline
\end{tabular}
\end{minipage}
\vspace{2mm}
\caption{Quantitative comparisons on CUFED5.}
\label{table:cufed5 comparison}
\end{table}

\begin{table}[t]
\scriptsize
\begin{minipage}[b]{38mm}
\begin{tabular}{@{\hspace{0.2 mm}}l@{\hspace{7 mm}}c@{\hspace{4 mm}}c@{\hspace{1 mm}}c@{}}
\hline
\textbf{SISR}  & PSNR & SSIM   \\
\hline
Bicubic&33.20 & 0.893 \\
RSRGAN~\cite{zhang2019ranksrgan}& 33.51 & 0.873\\
RCAN~\cite{zhang2018rcan} &33.94 & 0.911\\
CSNLN~\cite{Mei2020image}& 36.10 & 0.927\\
\hline
\end{tabular}

\end{minipage}
\qquad
\begin{minipage}[b]{38mm}
\begin{tabular}{@{\hspace{0.2 mm}}l@{\hspace{7 mm}}c@{\hspace{4 mm}}c@{\hspace{1 mm}}c@{}}
\hline
\textbf{RefSR}  & PSNR & SSIM  \\
\hline
TTSR ~\cite{yang2020learning}&35.48 & 0.915\\
TTSR-\textit{$\ell_1$}~\cite{yang2020learning} & 36.28 & 0.928\\
Ours &34.41 & 0.904\\
 \textbf{Ours-\textit{$\ell_1$}}&\textbf{36.98} & \textbf{0.933} \\
\hline
\end{tabular}
\end{minipage}

\vspace{2mm}
\caption{Quantitative comparison on the CameraFusion dataset.}
\label{table:st comparison}
\end{table}

\begin{figure}[t]
    \centering
    \includegraphics[width=0.9\linewidth]{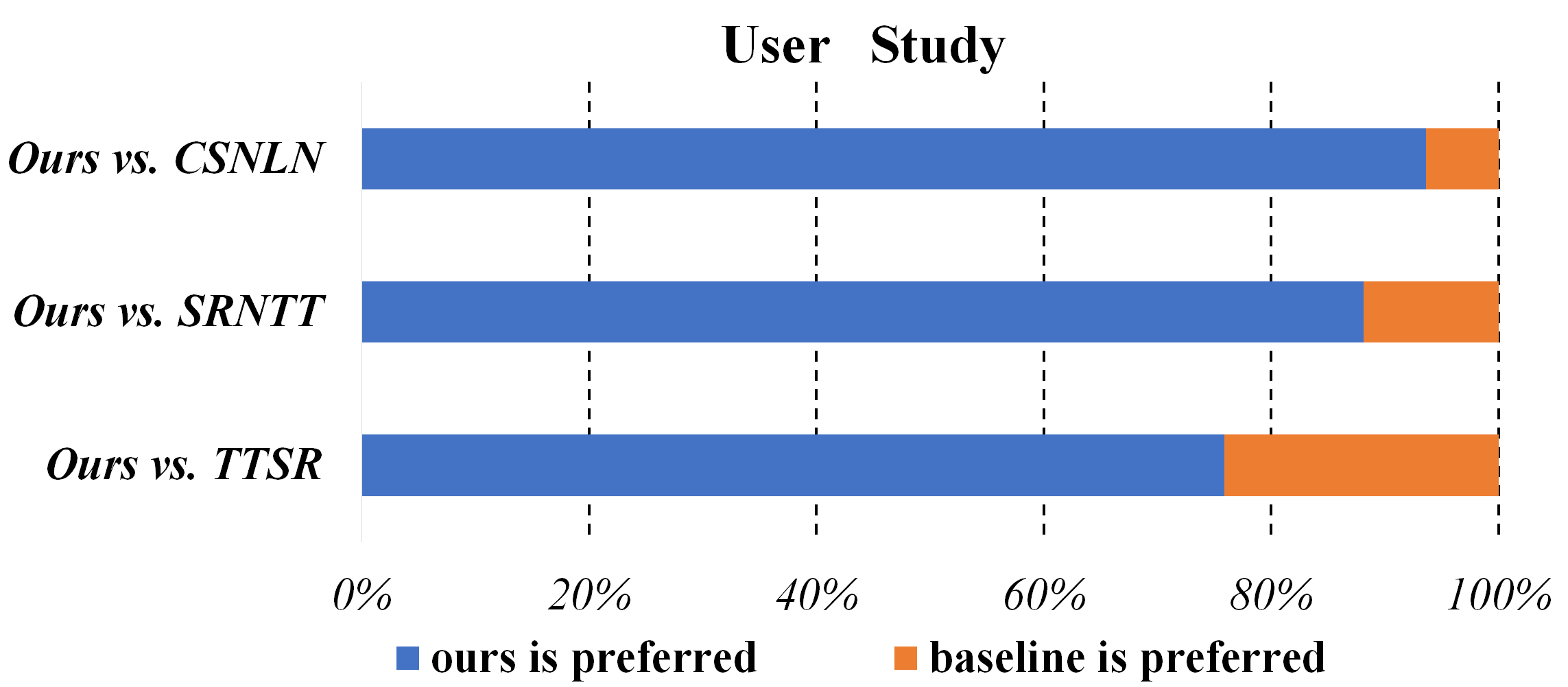}
    \caption{User study results on CUFED5. The reported values indicate the preference rate of our results against other approaches.}
    \label{fig:user}
\end{figure}
\subsection{Evaluation }
\subsubsection{Evaluation on CUFED5}

\begin{figure*}[t]
\centering
\scalebox{1.05}{
\begin{tabular}{@{}c@{\hspace{0.5mm}}c@{\hspace{1.5mm}}c@{\hspace{0.5mm}}c@{\hspace{1.5mm}}c@{\hspace{0.5mm}}c@{\hspace{1.5mm}}c@{\hspace{0.5mm}}c@{}}
\includegraphics[width=0.15\linewidth,height=0.08\linewidth]{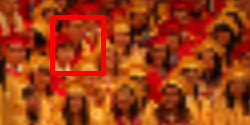}&
\includegraphics[width=0.07\linewidth,height=0.08\linewidth]{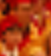}&
\includegraphics[width=0.15\linewidth,height=0.08\linewidth]{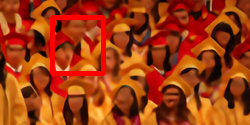}&
\includegraphics[width=0.07\linewidth,height=0.08\linewidth]{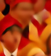}&
\includegraphics[width=0.15\linewidth,height=0.08\linewidth]{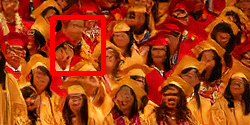}&
\includegraphics[width=0.07\linewidth,height=0.08\linewidth]{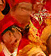}&
\includegraphics[width=0.15\linewidth,height=0.08\linewidth]{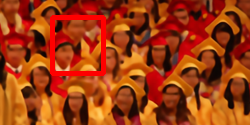}&
\includegraphics[width=0.07\linewidth,height=0.08\linewidth]{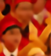}\\
\small Input&&\small RCAN~\cite{zhang2018rcan}&&\small RSRGAN~\cite{zhang2019ranksrgan}&&\small CSNLN~\cite{Mei2020image}&\\

\includegraphics[width=0.15\linewidth,height=0.08\linewidth]{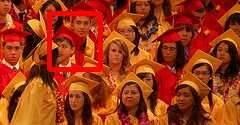}&
\includegraphics[width=0.07\linewidth,height=0.08\linewidth]{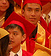}&
\includegraphics[width=0.15\linewidth,height=0.08\linewidth]{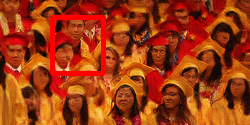}&
\includegraphics[width=0.07\linewidth,height=0.08\linewidth]{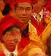}&
\includegraphics[width=0.15\linewidth,height=0.08\linewidth]{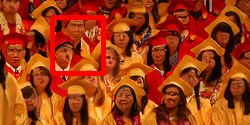}&
\includegraphics[width=0.07\linewidth,height=0.08\linewidth]{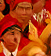}&
\includegraphics[width=0.15\linewidth,height=0.08\linewidth]{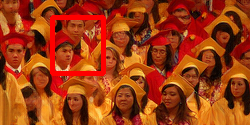}&
\includegraphics[width=0.07\linewidth,height=0.08\linewidth]{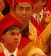}\\
\small Reference&&\small SRNTT~\cite{zhang2019image}&&\small TTSR~\cite{yang2020learning}&&\small Ours&\\

\includegraphics[width=0.15\linewidth,height=0.08\linewidth]{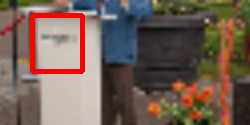}&
\includegraphics[width=0.07\linewidth,height=0.08\linewidth]{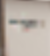}&
\includegraphics[width=0.15\linewidth,height=0.08\linewidth]{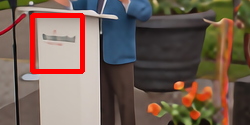}&
\includegraphics[width=0.07\linewidth,height=0.08\linewidth]{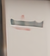}&
\includegraphics[width=0.15\linewidth,height=0.08\linewidth]{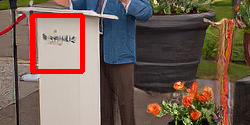}&
\includegraphics[width=0.07\linewidth,height=0.08\linewidth]{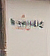}&
\includegraphics[width=0.15\linewidth,height=0.08\linewidth]{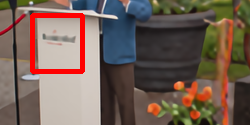}&
\includegraphics[width=0.07\linewidth,height=0.08\linewidth]{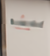}\\

\includegraphics[width=0.15\linewidth,height=0.08\linewidth]{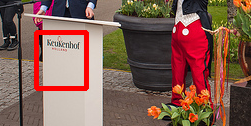}&
\includegraphics[width=0.07\linewidth,height=0.08\linewidth]{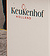}&
\includegraphics[width=0.15\linewidth,height=0.08\linewidth]{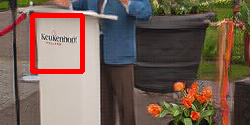}&
\includegraphics[width=0.07\linewidth,height=0.08\linewidth]{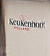}&
\includegraphics[width=0.15\linewidth,height=0.08\linewidth]{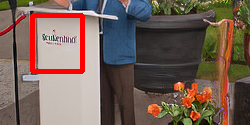}&
\includegraphics[width=0.07\linewidth,height=0.08\linewidth]{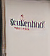}&
\includegraphics[width=0.15\linewidth,height=0.08\linewidth]{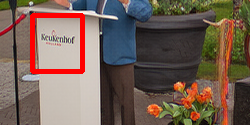}&
\includegraphics[width=0.07\linewidth,height=0.08\linewidth]{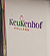}\\

\includegraphics[width=0.15\linewidth,height=0.08\linewidth]{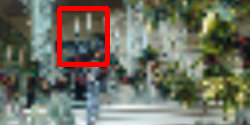}&
\includegraphics[width=0.07\linewidth,height=0.08\linewidth]{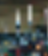}&
\includegraphics[width=0.15\linewidth,height=0.08\linewidth]{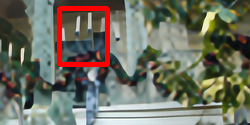}&
\includegraphics[width=0.07\linewidth,height=0.08\linewidth]{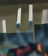}&
\includegraphics[width=0.15\linewidth,height=0.08\linewidth]{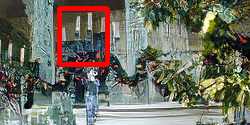}&
\includegraphics[width=0.07\linewidth,height=0.08\linewidth]{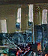}&
\includegraphics[width=0.15\linewidth,height=0.08\linewidth]{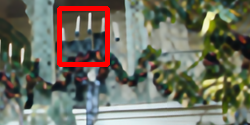}&
\includegraphics[width=0.07\linewidth,height=0.08\linewidth]{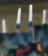}\\

\includegraphics[width=0.15\linewidth,height=0.08\linewidth]{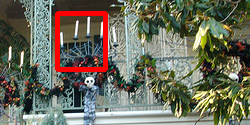}&
\includegraphics[width=0.07\linewidth,height=0.08\linewidth]{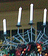}&
\includegraphics[width=0.15\linewidth,height=0.08\linewidth]{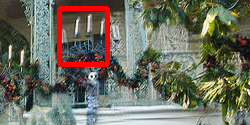}&
\includegraphics[width=0.07\linewidth,height=0.08\linewidth]{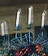}&
\includegraphics[width=0.15\linewidth,height=0.08\linewidth]{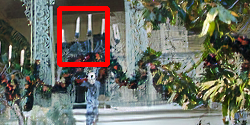}&
\includegraphics[width=0.07\linewidth,height=0.08\linewidth]{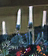}&
\includegraphics[width=0.15\linewidth,height=0.08\linewidth]{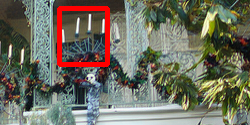}&
\includegraphics[width=0.07\linewidth,height=0.08\linewidth]{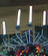}\\

\includegraphics[width=0.15\linewidth,height=0.08\linewidth]{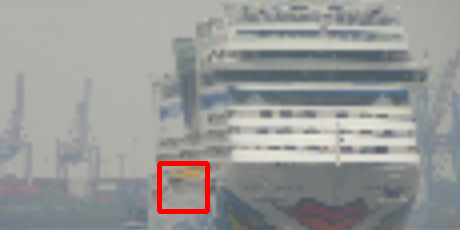}&
\includegraphics[width=0.07\linewidth,height=0.08\linewidth]{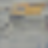}&
\includegraphics[width=0.15\linewidth,height=0.08\linewidth]{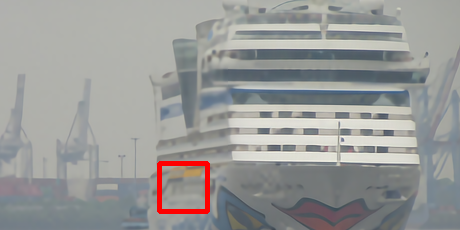}&
\includegraphics[width=0.07\linewidth,height=0.08\linewidth]{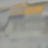}&
\includegraphics[width=0.15\linewidth,height=0.08\linewidth]{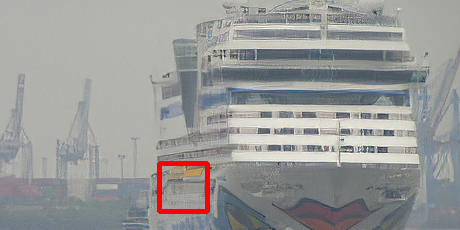}&
\includegraphics[width=0.07\linewidth,height=0.08\linewidth]{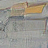}&
\includegraphics[width=0.15\linewidth,height=0.08\linewidth]{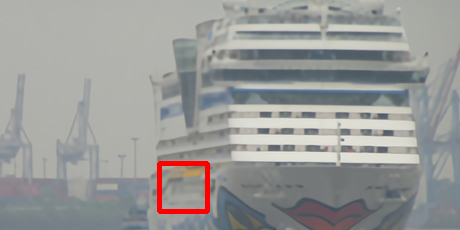}&
\includegraphics[width=0.07\linewidth,height=0.08\linewidth]{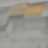}\\

\includegraphics[width=0.15\linewidth,height=0.08\linewidth]{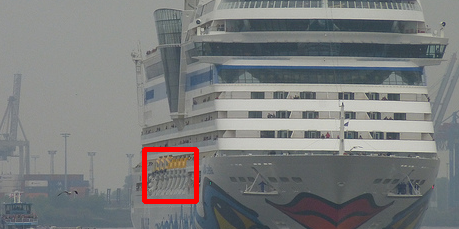}&
\includegraphics[width=0.07\linewidth,height=0.08\linewidth]{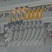}&
\includegraphics[width=0.15\linewidth,height=0.08\linewidth]{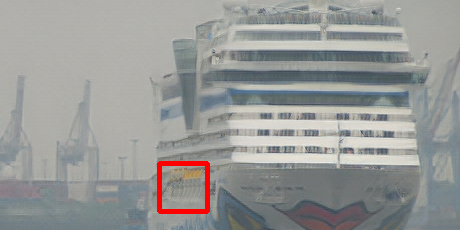}&
\includegraphics[width=0.07\linewidth,height=0.08\linewidth]{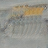}&
\includegraphics[width=0.15\linewidth,height=0.08\linewidth]{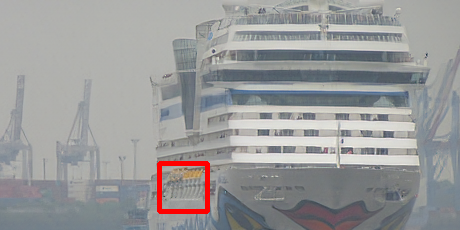}&
\includegraphics[width=0.07\linewidth,height=0.08\linewidth]{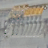}&
\includegraphics[width=0.15\linewidth,height=0.08\linewidth]{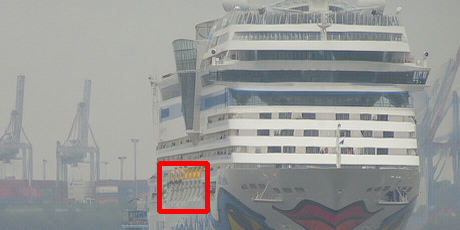}&
\includegraphics[width=0.07\linewidth,height=0.08\linewidth]{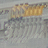}\\
\end{tabular}}
\caption{Qualitative comparisons on CUFED5. Our method reconstructs sharper and more realistic details than existing approaches for faces, texts, objects and textures. Zoom-in for details.}
\label{fig:CUFED5}
\end{figure*}
 
\quad \textbf{Quantitative Comparison} Table~\ref{table:cufed5 comparison} shows quantitative comparisons on CUFED5 in terms of PSNR and SSIM. It has been verified that due to trade-off between perception and distortion for super-resolution~\cite{wang2020deep},  visually-better results may suffer performance drop of PSNR. Therefore, we follow the setting in previous work~\cite{zhang2019image,yang2020learning} to re-train our model with $\ell_1$ loss only for fair comparison.
 
\textbf{Qualitative Comparison}  As shown in Fig.~\ref{fig:CUFED5}, our method shows better visual quality on faces, text, objects and textures. In the first example, human faces show with different orientations in the $I^{LR}$ and $I^{ Ref}$, while in the last example,  the cruise ship shows a larger size in $I^{ Ref}$ than $I^{LR}$ as it is moving forward to the camera. Despite the misalignment of orientations and scales, our model successfully obtains high-fidelity  results  via robust feature warping and fusion, while other methods either generate abrupt artifacts or produce blurry details. In other two examples, the scenes are statistic but have different view-points in $I^{LR}$ and $I^{ Ref}$, and we reconstruct recognizable texts and realistic textures.

\textbf{User Study}
We conduct a user study on Amazon Mechanical Turk (AMT) to compare our approach with state-of-the-art SISR~\cite{Mei2020image} and RefSR~\cite{zhang2019image, yang2020learning} methods. In specific, we provide participants with two images (ours and baselines) each time and ask them to select a more realistic one. We totally collect 1,920 valid votes from 16 participants. As shown in Fig.~\ref{fig:user}, we  outperforms previous work by a large margin.

\begin{figure*}[t]
\centering
\small
\begin{tabular}{@{}c@{\hspace{0.4mm}}c@{\hspace{0.4mm}}c@{\hspace{0.4mm}}c@{\hspace{0.4mm}}c@{\hspace{0.4mm}}c@{\hspace{0.4mm}}c@{}}
 Input (top) / Reference (bottom) &Bicubic &RCAN~\cite{zhang2018rcan}&RSRGAN~\cite{zhang2019ranksrgan}&CSNLN~\cite{Mei2020image}&TTSR~\cite{yang2020learning}&Ours\\
\toprule
\multirow{2}[2]{0.25\linewidth}[17mm]{ \includegraphics[width=\linewidth]{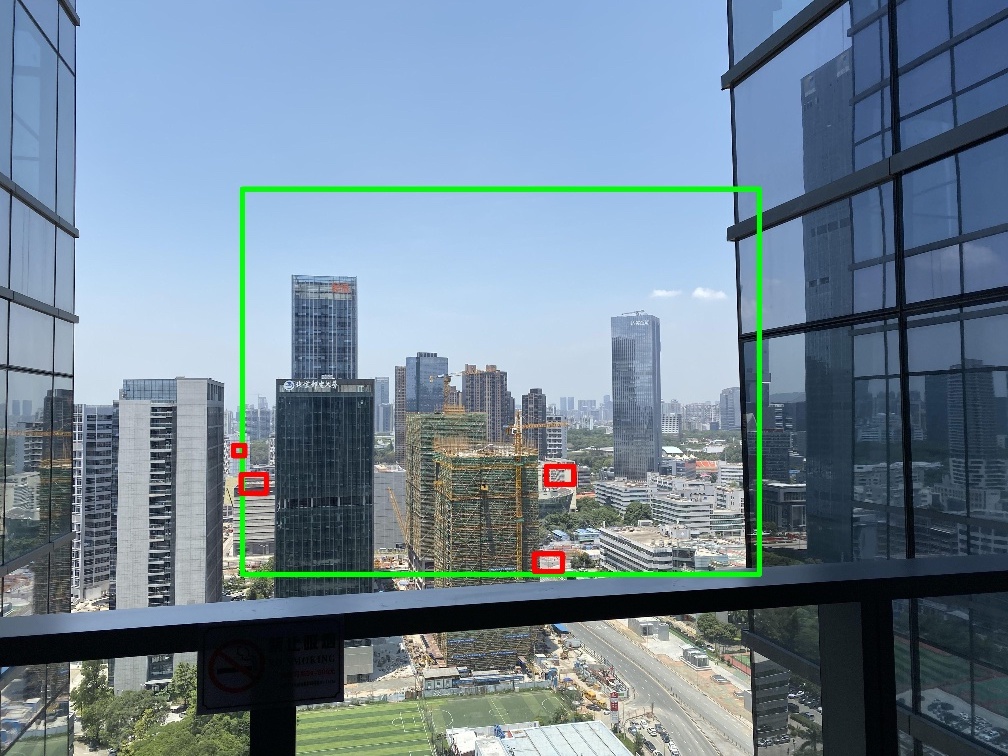}}&
\includegraphics[width=0.12\linewidth,height=0.09\linewidth]{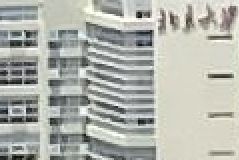}&
\includegraphics[width=0.12\linewidth,height=0.09\linewidth]{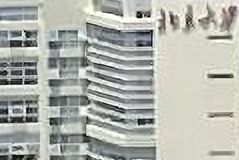}&
\includegraphics[width=0.12\linewidth,height=0.09\linewidth]{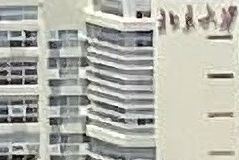}&
\includegraphics[width=0.12\linewidth,height=0.09\linewidth]{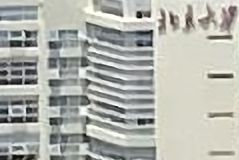}&
\includegraphics[width=0.12\linewidth,height=0.09\linewidth]{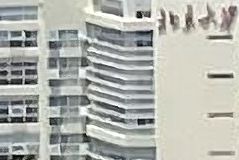}&
\includegraphics[width=0.12\linewidth,height=0.09\linewidth]{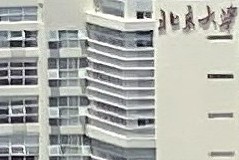}\\
&
\includegraphics[width=0.12\linewidth,height=0.09\linewidth]{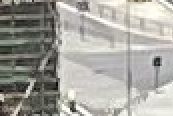}&
\includegraphics[width=0.12\linewidth,height=0.09\linewidth]{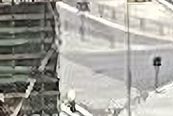}&
\includegraphics[width=0.12\linewidth,height=0.09\linewidth]{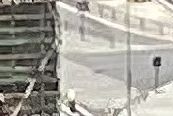}&
\includegraphics[width=0.12\linewidth,height=0.09\linewidth]{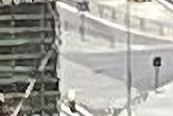}&
\includegraphics[width=0.12\linewidth,height=0.09\linewidth]{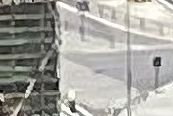}&
\includegraphics[width=0.12\linewidth,height=0.09\linewidth]{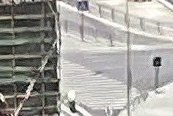}\\

\multirow{2}[2]{0.25\linewidth}[17mm]{ \includegraphics[width=\linewidth]{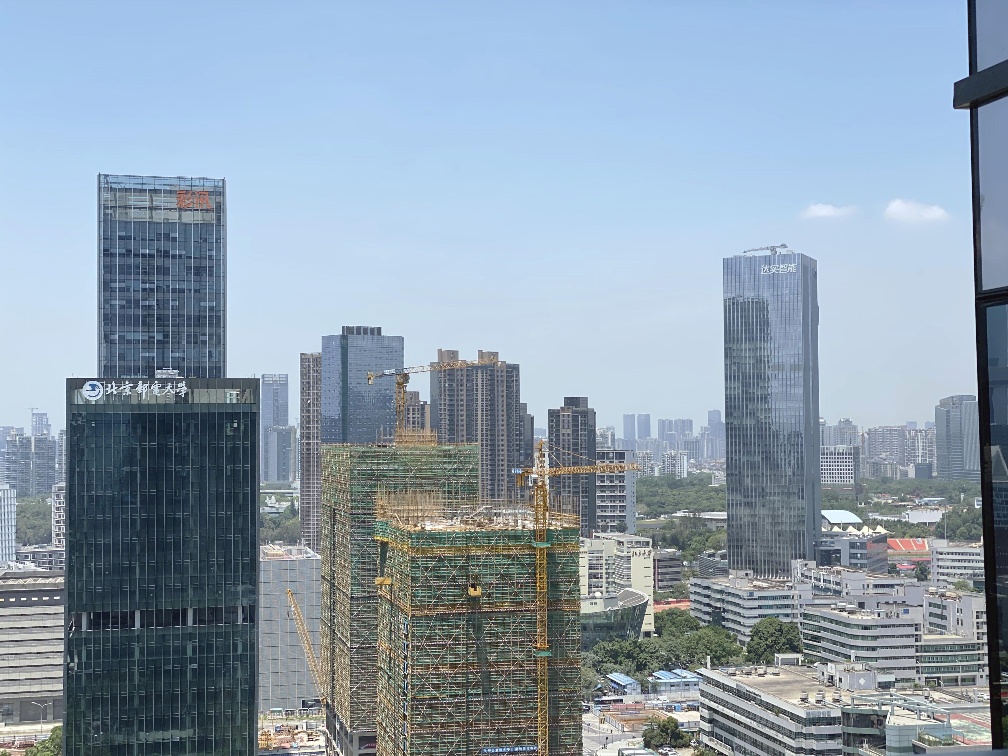}}&
\includegraphics[width=0.12\linewidth,height=0.09\linewidth]{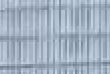}&
\includegraphics[width=0.12\linewidth,height=0.09\linewidth]{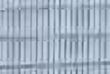}&
\includegraphics[width=0.12\linewidth,height=0.09\linewidth]{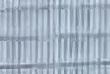}&
\includegraphics[width=0.12\linewidth,height=0.09\linewidth]{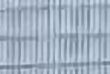}&
\includegraphics[width=0.12\linewidth,height=0.09\linewidth]{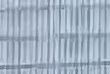}&
\includegraphics[width=0.12\linewidth,height=0.09\linewidth]{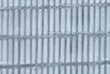}\\
&
\includegraphics[width=0.12\linewidth,height=0.09\linewidth]{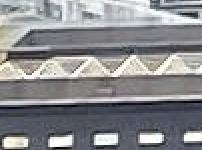}&
\includegraphics[width=0.12\linewidth,height=0.09\linewidth]{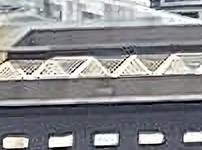}&
\includegraphics[width=0.12\linewidth,height=0.09\linewidth]{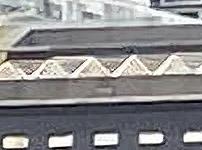}&
\includegraphics[width=0.12\linewidth,height=0.09\linewidth]{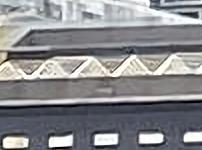}&
\includegraphics[width=0.12\linewidth,height=0.09\linewidth]{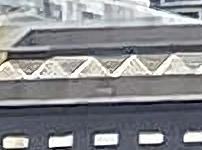}&
\includegraphics[width=0.12\linewidth,height=0.09\linewidth]{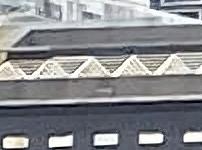}\\

\midrule

\multirow{2}[2]{0.25\linewidth}[17mm]{ \includegraphics[width=\linewidth]{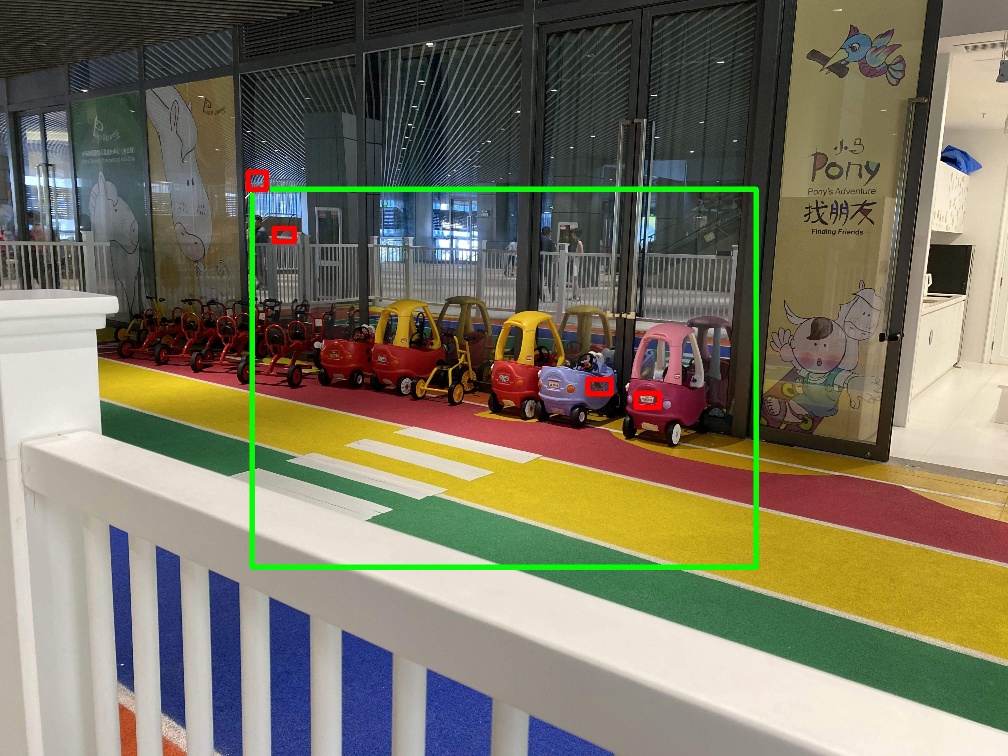}}&
\includegraphics[width=0.12\linewidth,height=0.09\linewidth]{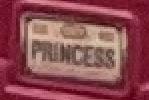}&
\includegraphics[width=0.12\linewidth,height=0.09\linewidth]{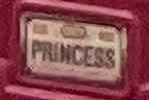}&
\includegraphics[width=0.12\linewidth,height=0.09\linewidth]{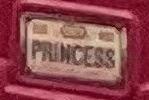}&
\includegraphics[width=0.12\linewidth,height=0.09\linewidth]{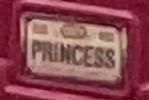}&
\includegraphics[width=0.12\linewidth,height=0.09\linewidth]{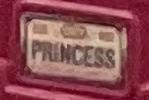}&
\includegraphics[width=0.12\linewidth,height=0.09\linewidth]{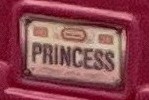}\\
&
\includegraphics[width=0.12\linewidth,height=0.09\linewidth]{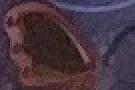}&
\includegraphics[width=0.12\linewidth,height=0.09\linewidth]{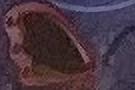}&
\includegraphics[width=0.12\linewidth,height=0.09\linewidth]{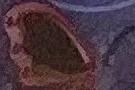}&
\includegraphics[width=0.12\linewidth,height=0.09\linewidth]{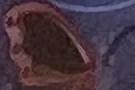}&
\includegraphics[width=0.12\linewidth,height=0.09\linewidth]{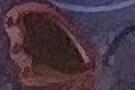}&
\includegraphics[width=0.12\linewidth,height=0.09\linewidth]{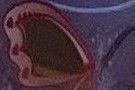}\\

\multirow{2}[2]{0.25\linewidth}[17mm]{ \includegraphics[width=\linewidth]{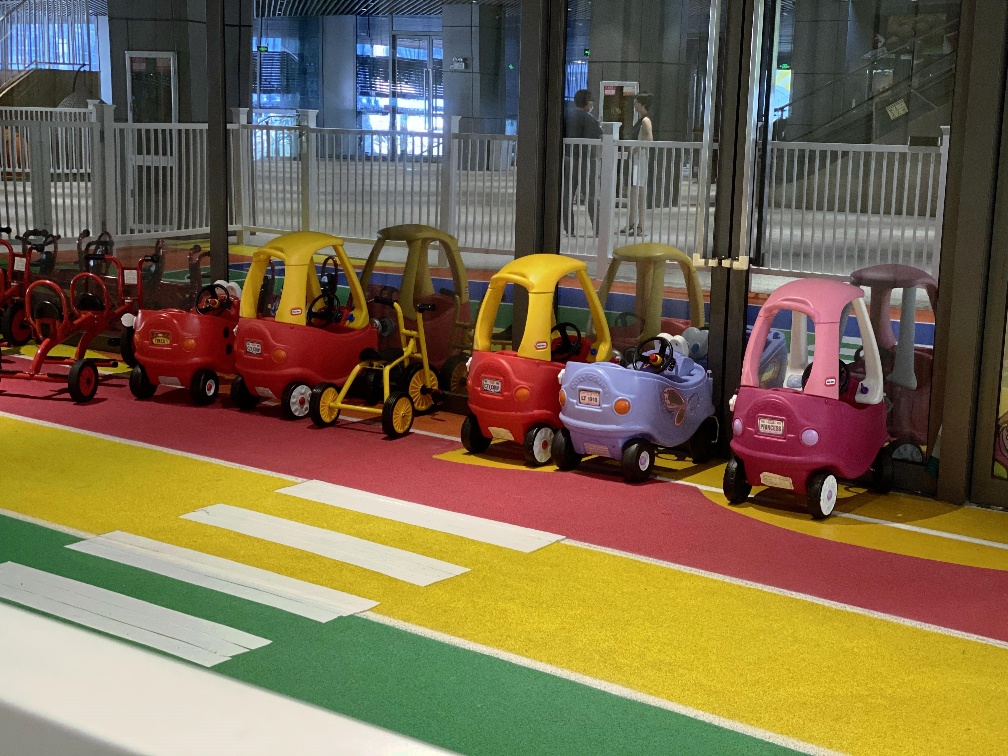}}&
\includegraphics[width=0.12\linewidth,height=0.09\linewidth]{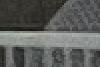}&
\includegraphics[width=0.12\linewidth,height=0.09\linewidth]{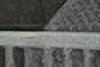}&
\includegraphics[width=0.12\linewidth,height=0.09\linewidth]{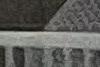}&
\includegraphics[width=0.12\linewidth,height=0.09\linewidth]{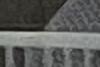}&
\includegraphics[width=0.12\linewidth,height=0.09\linewidth]{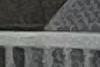}&
\includegraphics[width=0.12\linewidth,height=0.09\linewidth]{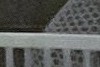}\\
&
\includegraphics[width=0.12\linewidth,height=0.09\linewidth]{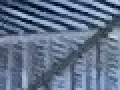}&
\includegraphics[width=0.12\linewidth,height=0.09\linewidth]{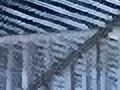}&
\includegraphics[width=0.12\linewidth,height=0.09\linewidth]{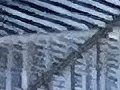}&
\includegraphics[width=0.12\linewidth,height=0.09\linewidth]{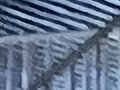}&
\includegraphics[width=0.12\linewidth,height=0.09\linewidth]{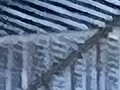}&
\includegraphics[width=0.12\linewidth,height=0.09\linewidth]{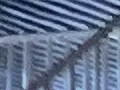}\\
\end{tabular}
\vspace{2mm}
\caption{Qualitative comparisons on the CameraFusion dataset. The green box indicates the overlapped FoV area between Input and Ref. Our method reconstructs sharper and more realistic details than previous approaches. Zoom-in for details. }
\label{fig:Our}
\end{figure*}

\subsubsection{Evaluation on CameraFusion}

To evaluate our method on dual-camera super-resolution, we re-train our model and  baselines on the  CameraFusion dataset. We select TTSR, CSNLN, RSRGAN and RCAN for comparison considering their outstanding performance on CUFED5. Specifically, we downsample  4K  wide-angle and  telephoto pairs to 2K-resolution   for training and metrics calculation in Table~\ref{table:st comparison}, by regarding 4K wide-angle images as ground truth.  We also observe that our performance gap between the overlapped FoV (37.28 / 0.942) and other regions (36.94 / 0.931) is small. This implies our approach is robust to reference image with different similarity levels.

For qualitative comparison, we fine-tune the trained models with full-resolution inputs, as mentioned in Section.~\ref{da}. When inference, we can super-resolve the 4K-resolution inputs to obtain 8K results. As in Fig.~\ref{fig:Our}, our approach correctly transfers correlated patterns  to reconstruct higher-fidelity outputs within overlapped FoV. It also  achieves comparable or better performance outside the overlapped FoV where corresponding reference patches are unavailable.

\subsection{Ablation Study}
\subsubsection{Effect of  reference similarity-levels}
To analyze how  the performance of our method is related  to the reference images, we conduct experiments  on reference images with different similarity levels in CUFED5~\cite{zhang2019image}. In Table~\ref{table:different similarity level}, L1 provides the most similar reference images, while L4 is the least relevant level. Our model  suffers little degradation when the similarity level decreases, which means that our method can robustly reconstruct images with  reference images of different similarity levels.

\begin{table}
\begin{center}
\scriptsize
\scalebox{1.08}{
\def\arraystretch{1.05}
\begin{tabular}{@{}c@{\hspace{3mm}}c@{\hspace{3mm}}c@{\hspace{3mm}}c@{\hspace{3mm}}c@{}}
\hline
Similarity level & L1&L2&L3&L4  \\
\hline
CrossNet~\cite{zheng2018crossnet}& 25.48 / .764&25.48 / .764&25.47 /. 763&25.46 / .763\\
SRNTT-\textit{$\ell_2$}~\cite{zhang2019image}& 26.15 / .781 &26.04 / .776& 25.98 / .775& 25.95 / .774\\
SSEN~\cite{shim2020robust}& 26.78 / .791 & 26.52 / .783 & 26.48 / .782 & 26.42 / .781  \\
TTSR-\textit{$\ell_1$}~\cite{yang2020learning} & 26.99 / .800& 26.74 / .791 & 26.64 / .788 & 26.58 / .787 \\

\hline
Ours-\textit{$\ell_1$} & \textbf{27.30} / \textbf{.807} & \textbf{26.92} / \textbf{.795} & \textbf{26.80} / \textbf{.791}& \textbf{26.70} / \textbf{.788}\\
\hline
\end{tabular}}
\end{center}
\caption{Ablation result on  the similarity level of reference images.  CUFED5 provides  four reference images for each LR image ranked by the similarity level, where L1 is the most relevant one.}
\label{table:different similarity level}
\end{table}

\subsubsection{Effect of Aligned Attention}
\begin{figure}
    \centering
    \scriptsize
    \begin{tabular}{@{}c@{\hspace{0.5 mm}} c@{\hspace{0.5mm}}  c@{}}
        \includegraphics[width=0.391\linewidth]{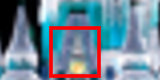} &  
         \includegraphics[width=0.277\linewidth]{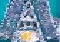} &
         \includegraphics[width=0.28\linewidth]{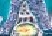} \\
        LR& (a) Patch Match&(b) Flow Alignment\\
        
        \includegraphics[width=0.391\linewidth]{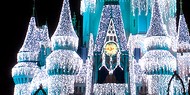} &  
        \includegraphics[width=0.277\linewidth]{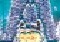} &  
        \includegraphics[width=0.28\linewidth]{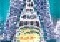}\\
        Reference&(c)  Ours (w/o alignment) &(d) Ours \\
    \end{tabular}
    \vspace{1mm}
    \caption{Ablation study on the aligned attention. The building presents different size and viewpoint in input and reference image, and we warp the reference by different methods.}
    \label{fig:aa}
\end{figure}

To further demonstrate how the  aligned attention facilitates the feature warping, we directly apply the feature-space index maps learned with and without the aligned attention to warp the original reference image. Note that since the index maps are originally learned to warp feature maps (instead of the images), the warped images are not the SR outputs and only used for visualization.   We also visualize the warped reference image  by flow-based method~\cite{zheng2018crossnet} and Patch Match~\cite{barnes2009patchmatch} for comparison. As shown in Fig.~\ref{fig:aa}, flow-based alignment leads to distorted structures, while Patch Match lacks high-quality details. In contrast, our model can  alleviate the spatial misalignment.

\begin{table}[t]
\small
\centering
\begin{tabular}{@{\hspace{2mm}}l@{\hspace{8mm}}c@{\hspace{5mm}}c@{\hspace{2mm}}}
\hline Feature Fusion Method  & PSNR & SSIM \\
\hline  
Element-wise Summation&   26.85 &  0.794  \\
Soft Fusion~\cite{yang2020learning} & 27.12 &  0.803   \\
Adaptive Fusion    & 27.30 &  0.807 \\
\hline
\end{tabular}
\vspace{2mm}
\caption{Ablation study on different feature fusion methods on  CUFED5.}
\label{tab: ablation}
\end{table}

\subsubsection{Effect of  Adaptive Fusion}
\begin{figure}
    \centering
    \scriptsize
    \begin{tabular}{@{}c@{\hspace{0.5 mm}} c@{\hspace{0.5mm}}   c@{}}
        \includegraphics[width=0.39\linewidth]{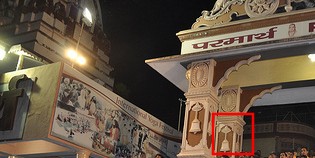} &  
         \includegraphics[width=0.29\linewidth]{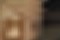} &  
        \includegraphics[width=0.29\linewidth]{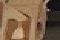} \\
        Ground Truth& (a) LR&(b) Soft Fusion\\
        
        \includegraphics[width=0.39\linewidth]{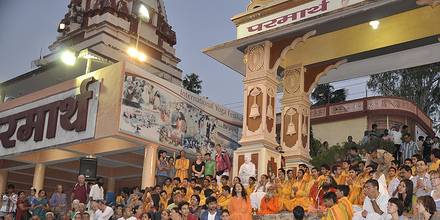} &  
        \includegraphics[width=0.29\linewidth]{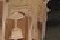}  &
        \includegraphics[width=0.29\linewidth]{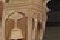} \\
        Reference&(c) Adaptive Fusion&(d) Adaptive Fusion \\
             &(feature only)&(feature + residual) \\
    \end{tabular}
    \vspace{1mm}
    \caption{Ablation study on the adaptive fusion. As shown in (c), with adaptive fusion only in the feature space, high-frequency details are not be fully  transferred.  }
    \label{fig:ff}
\end{figure}

Table~\ref{tab: ablation} provides ablation results on the adaptive fusion. We    apply element-wise summation (add $F_{aligned}^{ref}$ to $F^{SR}$ without confidence guidance), soft fusion~\cite{yang2020learning} (fuse $F_{aligned}^{ref}$ and $F^{SR}$ with original confidence map) and adaptive fusion (fuse $F_{aligned}^{ref}$ and $F^{SR}$ with learnable confidence map), respectively. With the adaptive fusion, we observe performance gain of 0.18 dB over the soft fusion, which implies the benefit from a learnable confidence map. As shown in Fig.~\ref{fig:ff}, by further applying adaptive fusion  for high-frequency residuals as Eq.~\ref{eq:image_sapce}, the model can generate sharper structures and more realistic textures.

\subsubsection{Effect of Fidelity Loss}
\begin{figure}
    \centering
    \scriptsize
    \begin{tabular}{@{}c@{\hspace{0.8mm}} c@{\hspace{0.5mm}} c@{\hspace{0.5mm}} c@{} }
         
        \multirow{2}[2]{0.452\linewidth}[11.2mm]{  \includegraphics[width=1\linewidth]{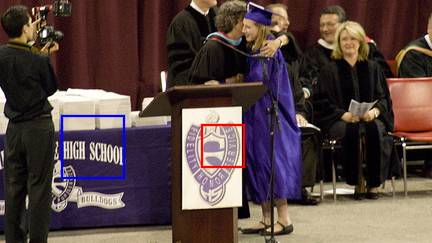}} & 
         \includegraphics[width=0.15\linewidth]{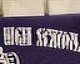} & 
         \includegraphics[width=0.15\linewidth]{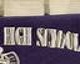}&
         \includegraphics[width=0.15\linewidth]{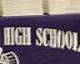}\\
         & 
        \includegraphics[width=0.15\linewidth]{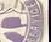} & 
         \includegraphics[width=0.15\linewidth]{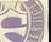}&
         \includegraphics[width=0.15\linewidth]{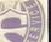}\\  
          Ground Truth&TTSR&Ours w/o &Ours with\\
         &&fidelity loss&fidelity loss\\

    \end{tabular}
    \vspace{1mm}
    \caption{Ablation experiment on the fidelity loss. With the fidelity loss, we can obtain higher-fidelity reconstruction results.}
    \label{fig:loss}
\end{figure}

The fidelity loss is imposed to enforce  the output SR image to possess high-quality details as reference images.  The key idea  is to adaptively maximize the similarity between Ref and SR  according to the matching confidence. Fig.~\ref{fig:loss} shows that without this loss,  the network fails to accurately utilize reference cues for high-fidelity generation, since LR features dominates the reconstruction process.

\begin{figure}[t]
\centering
\small
\begin{tabular}{@{}c@{\hspace{0.4mm}}c@{\hspace{0.4mm}}c@{\hspace{0.4mm}}}

\small Input&  Ours (w/o SRA) & Ours (w/ SRA)\\
 \multirow{2}[2]{0.48\linewidth}[16.2mm]{
 \includegraphics[width=1\linewidth,height=0.765\linewidth]{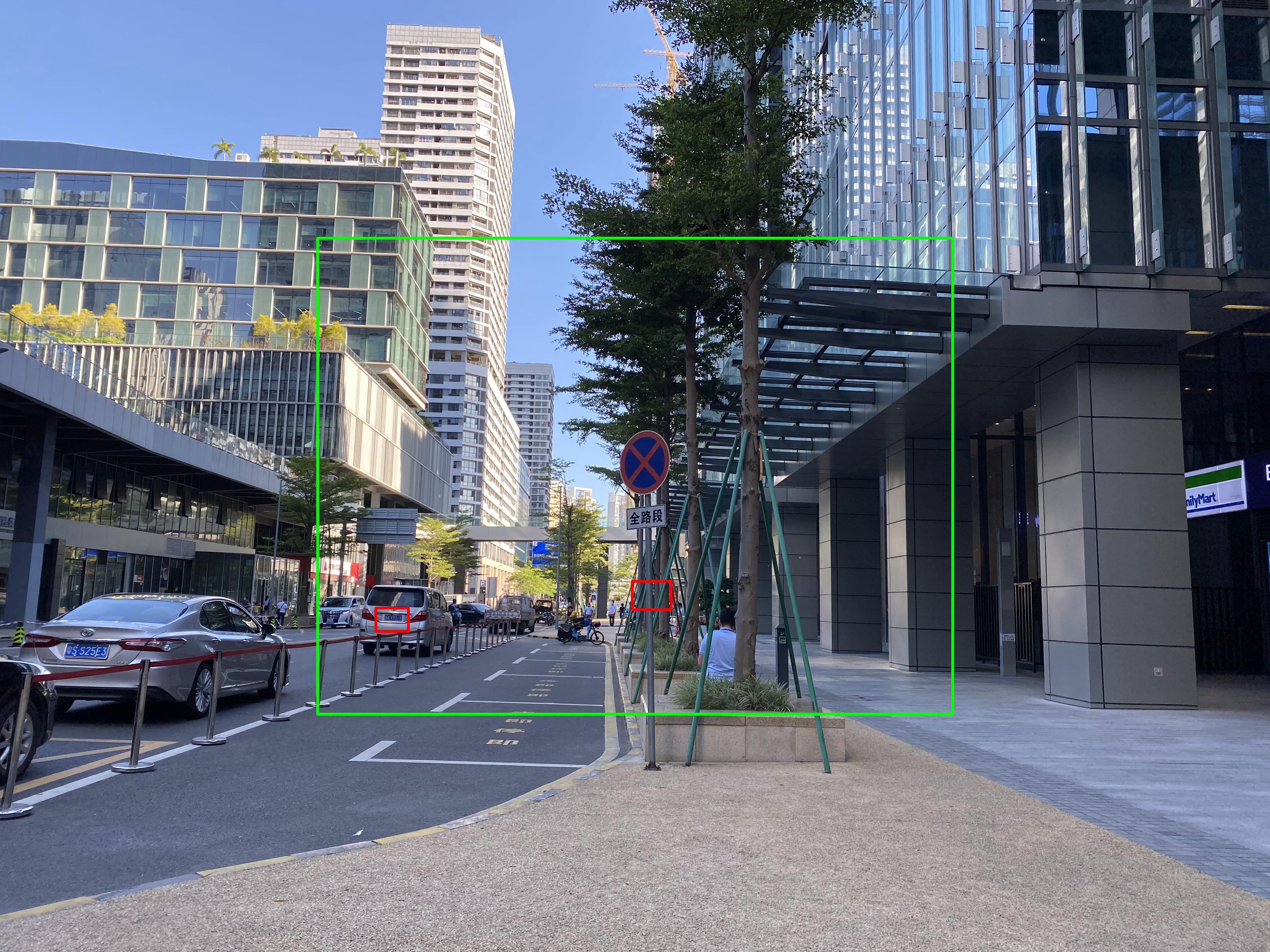}}&
\includegraphics[width=0.24\linewidth,height=0.178\linewidth]{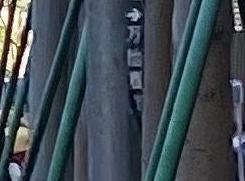}&
\includegraphics[width=0.24\linewidth,height=0.178\linewidth]{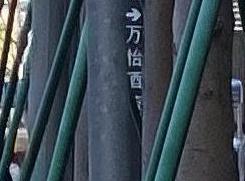}\\
&
\includegraphics[width=0.24\linewidth,height=0.178\linewidth]{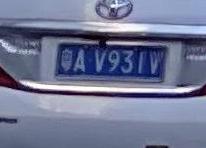}&
\includegraphics[width=0.24\linewidth,height=0.178\linewidth]{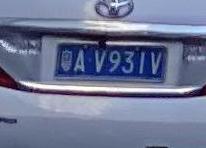}\\
 
\end{tabular}

\caption{Ablation study on SRA on CameraFusion dataset. Zoom-in for details. }
\label{fig:ft}
\end{figure}

\subsubsection{ Effect of Self-supervised Real-image Adaption }
As shwon  in Fig.~\ref{fig:ft}, without the proposed self-supervised real-image adaption, the super-resolution results on real-world camera photos are blurry.

\section{Conclusion }
In this paper, we study the reference-based super-resolution with the focus on real-world dual-camera zoom. To alleviate the spatial misalignment between input and reference images, we propose an aligned attention module for more robust feature warping. To advance our method to  dual-camera super-resolution for real-world smartphone images, we design a self-supervised domain adaptation scheme to generalize trained models to real-world inputs. Extensive experiments show that our method achieves compelling performance.

\section*{Acknowledgement}
This  project is supported by SenseTime Collaborative Research Grant. We thank Hao Ouyang and anonymous reviewers for helpful discussions and suggestions.

{\small
\bibliographystyle{ieee_fullname}
\bibliography{egbib}
}

\clearpage
\appendix 
 
\begin{figure*}[t!]
\centering
\begin{tabular}{@{}c@{\hspace{1mm}}c@{\hspace{1mm}}c@{\hspace{1mm}}c@{\hspace{1mm}}c@{\hspace{1mm}}c@{\hspace{1mm}}c@{\hspace{1mm}}c@{}}
\includegraphics[width=0.15\linewidth,height=0.08\linewidth]{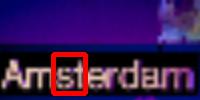}&
\includegraphics[width=0.07\linewidth,height=0.08\linewidth]{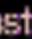}&
\includegraphics[width=0.15\linewidth,height=0.08\linewidth]{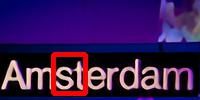}&
\includegraphics[width=0.07\linewidth,height=0.08\linewidth]{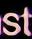}&
\includegraphics[width=0.15\linewidth,height=0.08\linewidth]{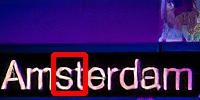}&
\includegraphics[width=0.07\linewidth,height=0.08\linewidth]{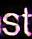}&
\includegraphics[width=0.15\linewidth,height=0.08\linewidth]{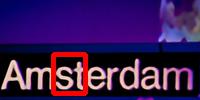}&
\includegraphics[width=0.07\linewidth,height=0.08\linewidth]{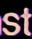}\\
\small Bicubic&&\small RCAN &&\small RSRGAN &&\small CSNLN &\\
\\
\includegraphics[width=0.15\linewidth,height=0.08\linewidth]{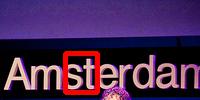}&
\includegraphics[width=0.07\linewidth,height=0.08\linewidth]{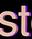}&
\includegraphics[width=0.15\linewidth,height=0.08\linewidth]{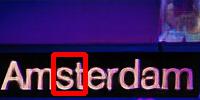}&
\includegraphics[width=0.07\linewidth,height=0.08\linewidth]{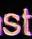}&
\includegraphics[width=0.15\linewidth,height=0.08\linewidth]{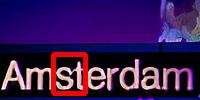}&
\includegraphics[width=0.07\linewidth,height=0.08\linewidth]{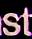}&
\includegraphics[width=0.15\linewidth,height=0.08\linewidth]{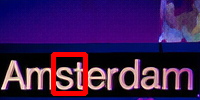}&
\includegraphics[width=0.07\linewidth,height=0.08\linewidth]{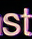}\\
\small Reference&&SRNTT &&TTSR &&Ours&\\

\\\hline\\
\\
\includegraphics[width=0.15\linewidth,height=0.08\linewidth]{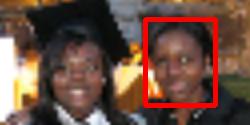}&
\includegraphics[width=0.07\linewidth,height=0.08\linewidth]{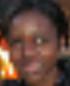}&
\includegraphics[width=0.15\linewidth,height=0.08\linewidth]{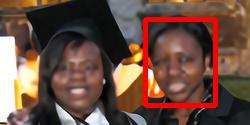}&
\includegraphics[width=0.07\linewidth,height=0.08\linewidth]{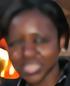}&
\includegraphics[width=0.15\linewidth,height=0.08\linewidth]{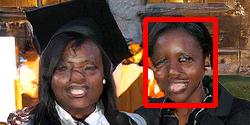}&
\includegraphics[width=0.07\linewidth,height=0.08\linewidth]{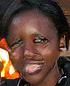}&
\includegraphics[width=0.15\linewidth,height=0.08\linewidth]{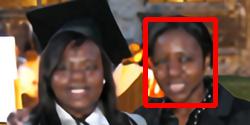}&
\includegraphics[width=0.07\linewidth,height=0.08\linewidth]{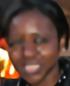}\\
\small Bicubic&&\small RCAN &&\small RSRGAN &&\small CSNLN &\\
\\
\includegraphics[width=0.15\linewidth,height=0.08\linewidth]{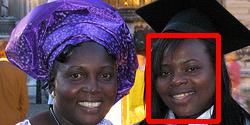}&
\includegraphics[width=0.07\linewidth,height=0.08\linewidth]{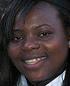}&
\includegraphics[width=0.15\linewidth,height=0.08\linewidth]{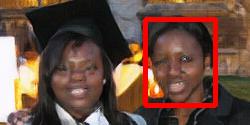}&
\includegraphics[width=0.07\linewidth,height=0.08\linewidth]{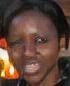}&
\includegraphics[width=0.15\linewidth,height=0.08\linewidth]{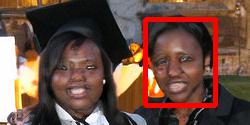}&
\includegraphics[width=0.07\linewidth,height=0.08\linewidth]{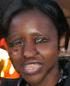}&
\includegraphics[width=0.15\linewidth,height=0.08\linewidth]{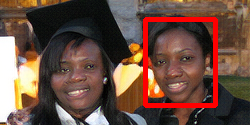}&
\includegraphics[width=0.07\linewidth,height=0.08\linewidth]{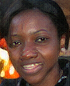}\\
\small Reference&&SRNTT &&TTSR &&Ours&\\

\\\hline\\
\\
\includegraphics[width=0.15\linewidth,height=0.08\linewidth]{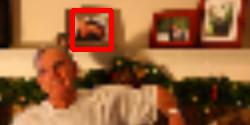}&
\includegraphics[width=0.07\linewidth,height=0.08\linewidth]{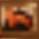}&
\includegraphics[width=0.15\linewidth,height=0.08\linewidth]{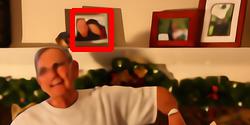}&
\includegraphics[width=0.07\linewidth,height=0.08\linewidth]{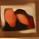}&
\includegraphics[width=0.15\linewidth,height=0.08\linewidth]{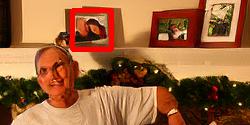}&
\includegraphics[width=0.07\linewidth,height=0.08\linewidth]{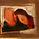}&
\includegraphics[width=0.15\linewidth,height=0.08\linewidth]{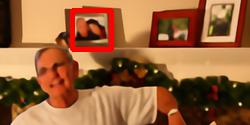}&
\includegraphics[width=0.07\linewidth,height=0.08\linewidth]{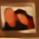}\\
\small Bicubic&&\small RCAN &&\small RSRGAN &&\small CSNLN &\\
\\
\includegraphics[width=0.15\linewidth,height=0.08\linewidth]{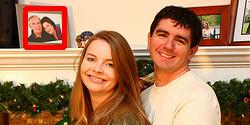}&
\includegraphics[width=0.07\linewidth,height=0.08\linewidth]{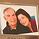}&
\includegraphics[width=0.15\linewidth,height=0.08\linewidth]{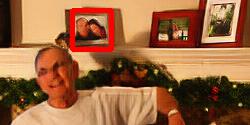}&
\includegraphics[width=0.07\linewidth,height=0.08\linewidth]{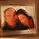}&
\includegraphics[width=0.15\linewidth,height=0.08\linewidth]{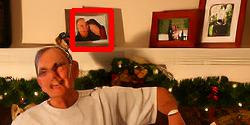}&
\includegraphics[width=0.07\linewidth,height=0.08\linewidth]{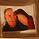}&
\includegraphics[width=0.15\linewidth,height=0.08\linewidth]{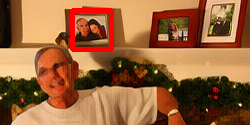}&
\includegraphics[width=0.07\linewidth,height=0.08\linewidth]{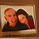}\\
\small Reference&&SRNTT &&TTSR &&Ours&\\

\\\hline\\
\\
\includegraphics[width=0.15\linewidth,height=0.08\linewidth]{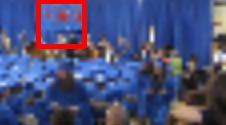}&
\includegraphics[width=0.07\linewidth,height=0.08\linewidth]{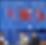}&
\includegraphics[width=0.15\linewidth,height=0.08\linewidth]{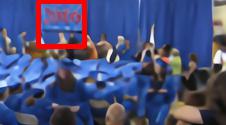}&
\includegraphics[width=0.07\linewidth,height=0.08\linewidth]{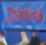}&
\includegraphics[width=0.15\linewidth,height=0.08\linewidth]{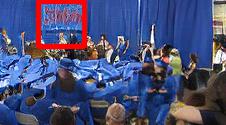}&
\includegraphics[width=0.07\linewidth,height=0.08\linewidth]{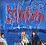}&
\includegraphics[width=0.15\linewidth,height=0.08\linewidth]{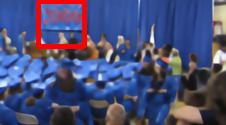}&
\includegraphics[width=0.07\linewidth,height=0.08\linewidth]{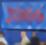}\\
\small Bicubic&&\small RCAN &&\small RSRGAN &&\small CSNLN &\\
\\
\includegraphics[width=0.15\linewidth,height=0.08\linewidth]{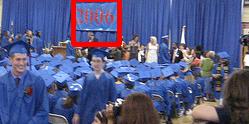}&
\includegraphics[width=0.07\linewidth,height=0.08\linewidth]{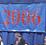}&
\includegraphics[width=0.15\linewidth,height=0.08\linewidth]{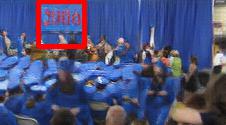}&
\includegraphics[width=0.07\linewidth,height=0.08\linewidth]{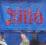}&
\includegraphics[width=0.15\linewidth,height=0.08\linewidth]{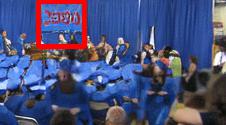}&
\includegraphics[width=0.07\linewidth,height=0.08\linewidth]{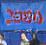}&
\includegraphics[width=0.15\linewidth,height=0.08\linewidth]{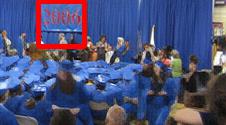}&
\includegraphics[width=0.07\linewidth,height=0.08\linewidth]{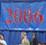}\\
\small Reference&&SRNTT &&TTSR &&Ours&\\

\end{tabular}
\vspace{2mm}
\caption{More qualitative comparisons on the CUFED5 dataset.}
\label{fig:CUFED5_supply}
\end{figure*}

\begin{figure*}[t!]
\centering
\begin{tabular}{@{}c@{\hspace{1mm}}c@{\hspace{1mm}}c@{\hspace{1mm}}c@{\hspace{1mm}}c@{\hspace{1mm}}c@{\hspace{1mm}}c@{\hspace{1mm}}c@{}}

\includegraphics[width=0.15\linewidth,height=0.08\linewidth]{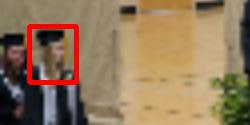}&
\includegraphics[width=0.07\linewidth,height=0.08\linewidth]{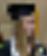}&
\includegraphics[width=0.15\linewidth,height=0.08\linewidth]{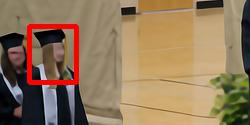}&
\includegraphics[width=0.07\linewidth,height=0.08\linewidth]{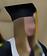}&
\includegraphics[width=0.15\linewidth,height=0.08\linewidth]{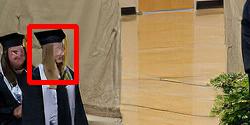}&
\includegraphics[width=0.07\linewidth,height=0.08\linewidth]{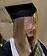}&
\includegraphics[width=0.15\linewidth,height=0.08\linewidth]{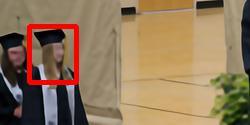}&
\includegraphics[width=0.07\linewidth,height=0.08\linewidth]{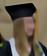}\\
\small Bicubic&&\small RCAN &&\small RSRGAN &&\small CSNLN &\\

\includegraphics[width=0.15\linewidth,height=0.08\linewidth]{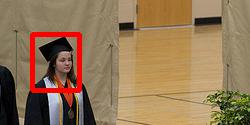}&
\includegraphics[width=0.07\linewidth,height=0.08\linewidth]{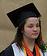}&
\includegraphics[width=0.15\linewidth,height=0.08\linewidth]{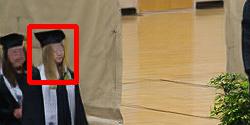}&
\includegraphics[width=0.07\linewidth,height=0.08\linewidth]{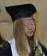}&
\includegraphics[width=0.15\linewidth,height=0.08\linewidth]{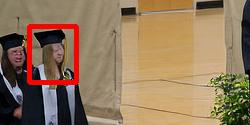}&
\includegraphics[width=0.07\linewidth,height=0.08\linewidth]{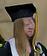}&
\includegraphics[width=0.15\linewidth,height=0.08\linewidth]{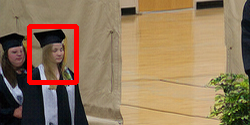}&
\includegraphics[width=0.07\linewidth,height=0.08\linewidth]{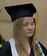}\\
\small Reference&&SRNTT &&TTSR &&Ours&\\
\\
\hline
\\

\includegraphics[width=0.15\linewidth,height=0.08\linewidth]{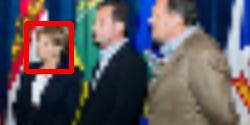}&
\includegraphics[width=0.07\linewidth,height=0.08\linewidth]{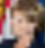}&
\includegraphics[width=0.15\linewidth,height=0.08\linewidth]{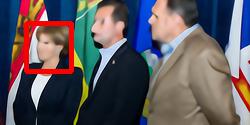}&
\includegraphics[width=0.07\linewidth,height=0.08\linewidth]{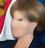}&
\includegraphics[width=0.15\linewidth,height=0.08\linewidth]{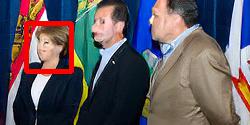}&
\includegraphics[width=0.07\linewidth,height=0.08\linewidth]{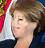}&
\includegraphics[width=0.15\linewidth,height=0.08\linewidth]{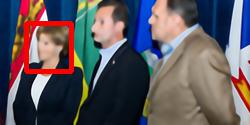}&
\includegraphics[width=0.07\linewidth,height=0.08\linewidth]{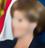}\\
\small Bicubic&&\small RCAN &&\small RSRGAN &&\small CSNLN &\\
\includegraphics[width=0.15\linewidth,height=0.08\linewidth]{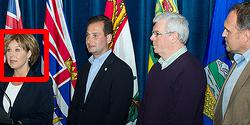}&
\includegraphics[width=0.07\linewidth,height=0.08\linewidth]{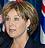}&
\includegraphics[width=0.15\linewidth,height=0.08\linewidth]{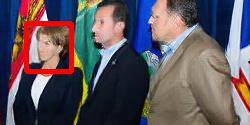}&
\includegraphics[width=0.07\linewidth,height=0.08\linewidth]{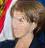}&
\includegraphics[width=0.15\linewidth,height=0.08\linewidth]{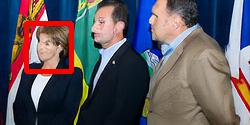}&
\includegraphics[width=0.07\linewidth,height=0.08\linewidth]{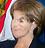}&
\includegraphics[width=0.15\linewidth,height=0.08\linewidth]{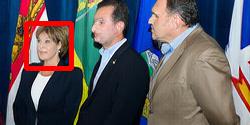}&
\includegraphics[width=0.07\linewidth,height=0.08\linewidth]{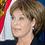}\\
\small Reference&&SRNTT &&TTSR &&Ours&\\
\\
\hline
\\
 
\includegraphics[width=0.15\linewidth,height=0.08\linewidth]{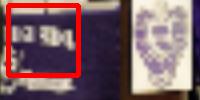}&
\includegraphics[width=0.07\linewidth,height=0.08\linewidth]{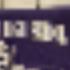}&
\includegraphics[width=0.15\linewidth,height=0.08\linewidth]{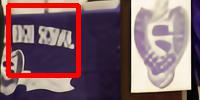}&
\includegraphics[width=0.07\linewidth,height=0.08\linewidth]{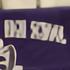}&
\includegraphics[width=0.15\linewidth,height=0.08\linewidth]{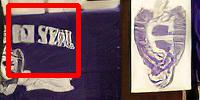}&
\includegraphics[width=0.07\linewidth,height=0.08\linewidth]{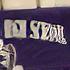}&
\includegraphics[width=0.15\linewidth,height=0.08\linewidth]{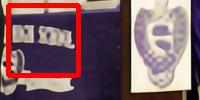}&
\includegraphics[width=0.07\linewidth,height=0.08\linewidth]{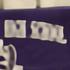}\\
\small Bicubic&&\small RCAN &&\small RSRGAN &&\small CSNLN &\\
\includegraphics[width=0.15\linewidth,height=0.08\linewidth]{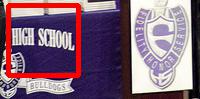}&
\includegraphics[width=0.07\linewidth,height=0.08\linewidth]{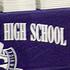}&
\includegraphics[width=0.15\linewidth,height=0.08\linewidth]{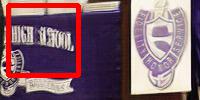}&
\includegraphics[width=0.07\linewidth,height=0.08\linewidth]{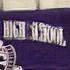}&
\includegraphics[width=0.15\linewidth,height=0.08\linewidth]{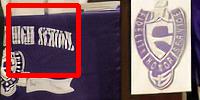}&
\includegraphics[width=0.07\linewidth,height=0.08\linewidth]{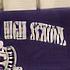}&
\includegraphics[width=0.15\linewidth,height=0.08\linewidth]{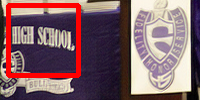}&
\includegraphics[width=0.07\linewidth,height=0.08\linewidth]{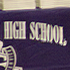}\\

\small Reference&&SRNTT &&TTSR &&Ours&\\
\\
\hline
\\
\includegraphics[width=0.15\linewidth,height=0.08\linewidth]{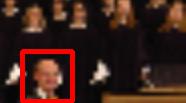}&
\includegraphics[width=0.07\linewidth,height=0.08\linewidth]{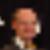}&
\includegraphics[width=0.15\linewidth,height=0.08\linewidth]{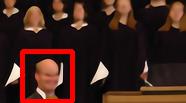}&
\includegraphics[width=0.07\linewidth,height=0.08\linewidth]{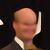}&
\includegraphics[width=0.15\linewidth,height=0.08\linewidth]{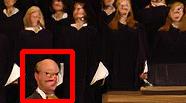}&
\includegraphics[width=0.07\linewidth,height=0.08\linewidth]{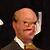}&
\includegraphics[width=0.15\linewidth,height=0.08\linewidth]{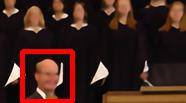}&
\includegraphics[width=0.07\linewidth,height=0.08\linewidth]{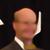}\\
\small Bicubic&&\small RCAN &&\small RSRGAN &&\small CSNLN &\\
\includegraphics[width=0.15\linewidth,height=0.08\linewidth]{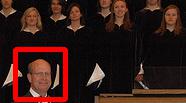}&
\includegraphics[width=0.07\linewidth,height=0.08\linewidth]{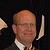}&
\includegraphics[width=0.15\linewidth,height=0.08\linewidth]{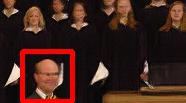}&
\includegraphics[width=0.07\linewidth,height=0.08\linewidth]{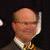}&
\includegraphics[width=0.15\linewidth,height=0.08\linewidth]{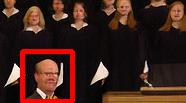}&
\includegraphics[width=0.07\linewidth,height=0.08\linewidth]{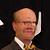}&
\includegraphics[width=0.15\linewidth,height=0.08\linewidth]{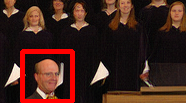}&
\includegraphics[width=0.07\linewidth,height=0.08\linewidth]{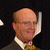}\\
\small Reference&&SRNTT &&TTSR &&Ours&\\
\\
\hline
\\
\includegraphics[width=0.15\linewidth,height=0.08\linewidth]{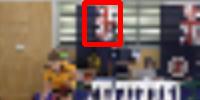}&
\includegraphics[width=0.07\linewidth,height=0.08\linewidth]{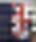}&
\includegraphics[width=0.15\linewidth,height=0.08\linewidth]{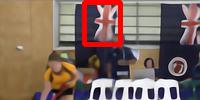}&
\includegraphics[width=0.07\linewidth,height=0.08\linewidth]{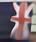}&
\includegraphics[width=0.15\linewidth,height=0.08\linewidth]{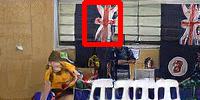}&
\includegraphics[width=0.07\linewidth,height=0.08\linewidth]{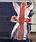}&
\includegraphics[width=0.15\linewidth,height=0.08\linewidth]{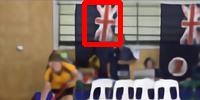}&
\includegraphics[width=0.07\linewidth,height=0.08\linewidth]{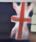}\\
\small Bicubic&&\small RCAN &&\small RSRGAN &&\small CSNLN &\\

\includegraphics[width=0.15\linewidth,height=0.08\linewidth]{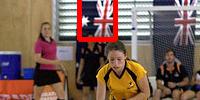}&
\includegraphics[width=0.07\linewidth,height=0.08\linewidth]{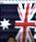}&
\includegraphics[width=0.15\linewidth,height=0.08\linewidth]{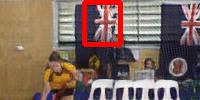}&
\includegraphics[width=0.07\linewidth,height=0.08\linewidth]{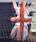}&
\includegraphics[width=0.15\linewidth,height=0.08\linewidth]{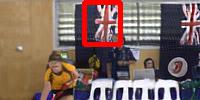}&
\includegraphics[width=0.07\linewidth,height=0.08\linewidth]{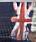}&
\includegraphics[width=0.15\linewidth,height=0.08\linewidth]{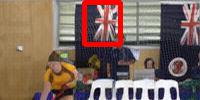}&
\includegraphics[width=0.07\linewidth,height=0.08\linewidth]{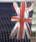}\\

\small Reference&&SRNTT &&TTSR &&Ours&\\

\end{tabular}
\vspace{2mm}
\caption{More qualitative comparisons on the CUFED5 dataset.}
\label{fig:CUFED5_supply_2}
\end{figure*}

\begin{figure*}[t!]
\centering
\begin{tabular}{@{}c@{\hspace{1mm}}c@{\hspace{1mm}}c@{\hspace{1mm}}c@{\hspace{1mm}}c@{\hspace{1mm}}c@{\hspace{1mm}}c@{\hspace{1mm}}c@{}}

\includegraphics[width=0.15\linewidth,height=0.08\linewidth]{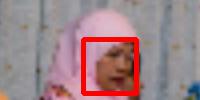}&
\includegraphics[width=0.07\linewidth,height=0.08\linewidth]{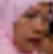}&
\includegraphics[width=0.15\linewidth,height=0.08\linewidth]{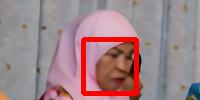}&
\includegraphics[width=0.07\linewidth,height=0.08\linewidth]{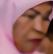}&
\includegraphics[width=0.15\linewidth,height=0.08\linewidth]{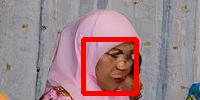}&
\includegraphics[width=0.07\linewidth,height=0.08\linewidth]{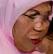}&
\includegraphics[width=0.15\linewidth,height=0.08\linewidth]{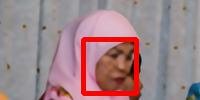}&
\includegraphics[width=0.07\linewidth,height=0.08\linewidth]{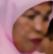}\\
\small Bicubic&&\small RCAN &&\small RSRGAN &&\small CSNLN &\\

\includegraphics[width=0.15\linewidth,height=0.08\linewidth]{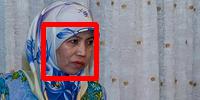}&
\includegraphics[width=0.07\linewidth,height=0.08\linewidth]{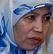}&
\includegraphics[width=0.15\linewidth,height=0.08\linewidth]{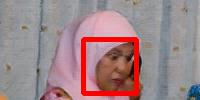}&
\includegraphics[width=0.07\linewidth,height=0.08\linewidth]{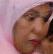}&
\includegraphics[width=0.15\linewidth,height=0.08\linewidth]{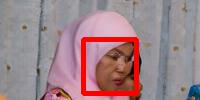}&
\includegraphics[width=0.07\linewidth,height=0.08\linewidth]{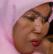}&
\includegraphics[width=0.15\linewidth,height=0.08\linewidth]{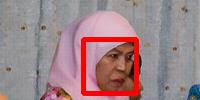}&
\includegraphics[width=0.07\linewidth,height=0.08\linewidth]{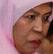}\\
\small Reference&&SRNTT &&TTSR &&Ours&\\
\\
\hline
\\
\includegraphics[width=0.15\linewidth,height=0.08\linewidth]{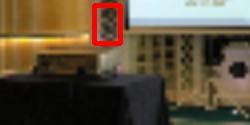}&
\includegraphics[width=0.07\linewidth,height=0.08\linewidth]{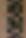}&
\includegraphics[width=0.15\linewidth,height=0.08\linewidth]{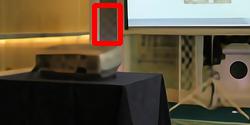}&
\includegraphics[width=0.07\linewidth,height=0.08\linewidth]{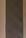}&
\includegraphics[width=0.15\linewidth,height=0.08\linewidth]{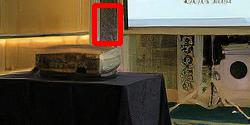}&
\includegraphics[width=0.07\linewidth,height=0.08\linewidth]{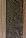}&
\includegraphics[width=0.15\linewidth,height=0.08\linewidth]{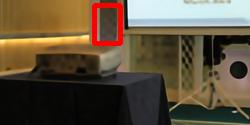}&
\includegraphics[width=0.07\linewidth,height=0.08\linewidth]{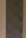}\\
\small Bicubic&&\small RCAN &&\small RSRGAN &&\small CSNLN &\\

\includegraphics[width=0.15\linewidth,height=0.08\linewidth]{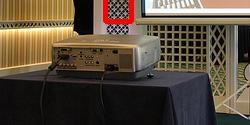}&
\includegraphics[width=0.07\linewidth,height=0.08\linewidth]{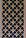}&
\includegraphics[width=0.15\linewidth,height=0.08\linewidth]{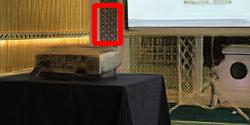}&
\includegraphics[width=0.07\linewidth,height=0.08\linewidth]{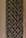}&
\includegraphics[width=0.15\linewidth,height=0.08\linewidth]{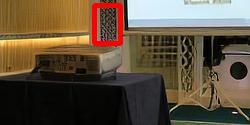}&
\includegraphics[width=0.07\linewidth,height=0.08\linewidth]{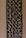}&
\includegraphics[width=0.15\linewidth,height=0.08\linewidth]{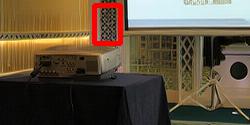}&
\includegraphics[width=0.07\linewidth,height=0.08\linewidth]{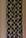}\\
\small Reference&&SRNTT &&TTSR &&Ours&\\
\\
\hline
\\

\includegraphics[width=0.15\linewidth,height=0.08\linewidth]{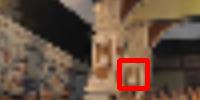}&
\includegraphics[width=0.07\linewidth,height=0.08\linewidth]{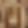}&
\includegraphics[width=0.15\linewidth,height=0.08\linewidth]{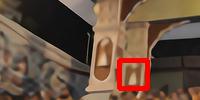}&
\includegraphics[width=0.07\linewidth,height=0.08\linewidth]{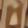}&
\includegraphics[width=0.15\linewidth,height=0.08\linewidth]{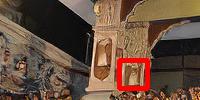}&
\includegraphics[width=0.07\linewidth,height=0.08\linewidth]{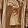}&
\includegraphics[width=0.15\linewidth,height=0.08\linewidth]{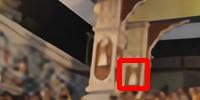}&
\includegraphics[width=0.07\linewidth,height=0.08\linewidth]{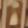}\\
\small Bicubic&&\small RCAN &&\small RSRGAN &&\small CSNLN &\\

\includegraphics[width=0.15\linewidth,height=0.08\linewidth]{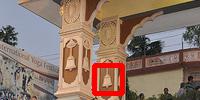}&
\includegraphics[width=0.07\linewidth,height=0.08\linewidth]{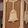}&
\includegraphics[width=0.15\linewidth,height=0.08\linewidth]{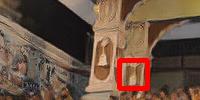}&
\includegraphics[width=0.07\linewidth,height=0.08\linewidth]{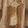}&
\includegraphics[width=0.15\linewidth,height=0.08\linewidth]{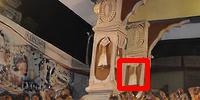}&
\includegraphics[width=0.07\linewidth,height=0.08\linewidth]{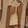}&
\includegraphics[width=0.15\linewidth,height=0.08\linewidth]{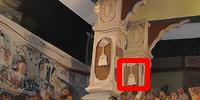}&
\includegraphics[width=0.07\linewidth,height=0.08\linewidth]{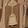}\\
\small Reference&&SRNTT &&TTSR  &&Ours&\\
\\
\hline
\\

\includegraphics[width=0.15\linewidth,height=0.08\linewidth]{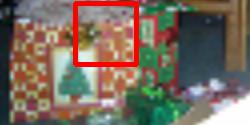}&
\includegraphics[width=0.07\linewidth,height=0.08\linewidth]{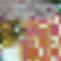}&
\includegraphics[width=0.15\linewidth,height=0.08\linewidth]{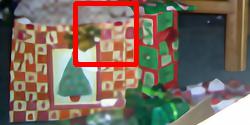}&
\includegraphics[width=0.07\linewidth,height=0.08\linewidth]{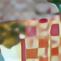}&
\includegraphics[width=0.15\linewidth,height=0.08\linewidth]{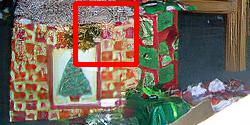}&
\includegraphics[width=0.07\linewidth,height=0.08\linewidth]{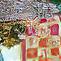}&
\includegraphics[width=0.15\linewidth,height=0.08\linewidth]{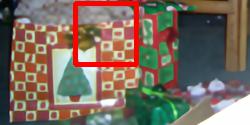}&
\includegraphics[width=0.07\linewidth,height=0.08\linewidth]{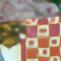}\\
\small Bicubic&&\small RCAN &&\small RSRGAN &&\small CSNLN &\\

\includegraphics[width=0.15\linewidth,height=0.08\linewidth]{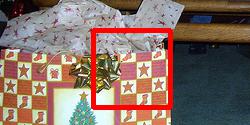}&
\includegraphics[width=0.07\linewidth,height=0.08\linewidth]{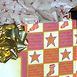}&
\includegraphics[width=0.15\linewidth,height=0.08\linewidth]{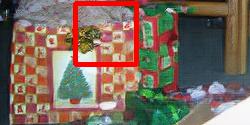}&
\includegraphics[width=0.07\linewidth,height=0.08\linewidth]{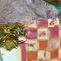}&
\includegraphics[width=0.15\linewidth,height=0.08\linewidth]{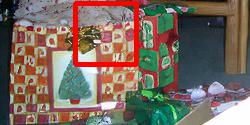}&
\includegraphics[width=0.07\linewidth,height=0.08\linewidth]{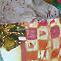}&
\includegraphics[width=0.15\linewidth,height=0.08\linewidth]{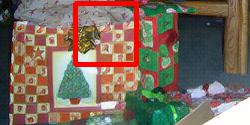}&
\includegraphics[width=0.07\linewidth,height=0.08\linewidth]{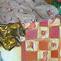}\\
\small Reference&&SRNTT &&TTSR &&Ours&\\
\\
\hline
\\

\includegraphics[width=0.15\linewidth,height=0.08\linewidth]{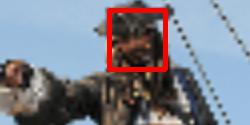}&
\includegraphics[width=0.07\linewidth,height=0.08\linewidth]{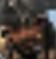}&
\includegraphics[width=0.15\linewidth,height=0.08\linewidth]{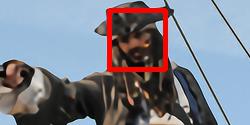}&
\includegraphics[width=0.07\linewidth,height=0.08\linewidth]{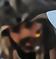}&
\includegraphics[width=0.15\linewidth,height=0.08\linewidth]{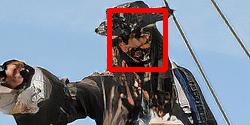}&
\includegraphics[width=0.07\linewidth,height=0.08\linewidth]{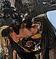}&
\includegraphics[width=0.15\linewidth,height=0.08\linewidth]{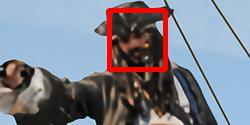}&
\includegraphics[width=0.07\linewidth,height=0.08\linewidth]{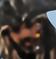}\\
\small Bicubic&&\small RCAN &&\small RSRGAN &&\small CSNLN &\\

\includegraphics[width=0.15\linewidth,height=0.08\linewidth]{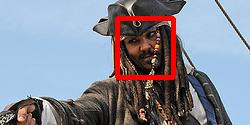}&
\includegraphics[width=0.07\linewidth,height=0.08\linewidth]{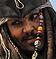}&
\includegraphics[width=0.15\linewidth,height=0.08\linewidth]{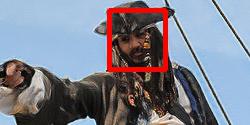}&
\includegraphics[width=0.07\linewidth,height=0.08\linewidth]{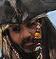}&
\includegraphics[width=0.15\linewidth,height=0.08\linewidth]{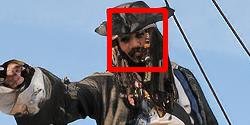}&
\includegraphics[width=0.07\linewidth,height=0.08\linewidth]{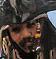}&
\includegraphics[width=0.15\linewidth,height=0.08\linewidth]{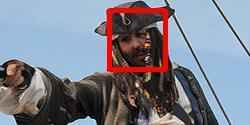}&
\includegraphics[width=0.07\linewidth,height=0.08\linewidth]{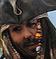}\\
\small Reference&&SRNTT &&TTSR &&Ours&\\

\end{tabular}
\vspace{2mm}
\caption{More qualitative comparisons on the CUFED5 dataset. }
\label{fig:CUFED5_supply_3}
\end{figure*}

\begin{figure*}[t]
\centering
\small
\begin{tabular}{@{}c@{\hspace{0.4mm}}c@{\hspace{0.4mm}}c@{\hspace{0.4mm}}c@{\hspace{0.4mm}}c@{\hspace{0.4mm}}c@{\hspace{0.4mm}}c@{}}
 Input  &Bicubic &RCAN &RSRGAN &CSNLN &TTSR &Ours\\

\multirow{2}[2]{0.17\linewidth}[20.5mm]{ \includegraphics[width=1\linewidth]{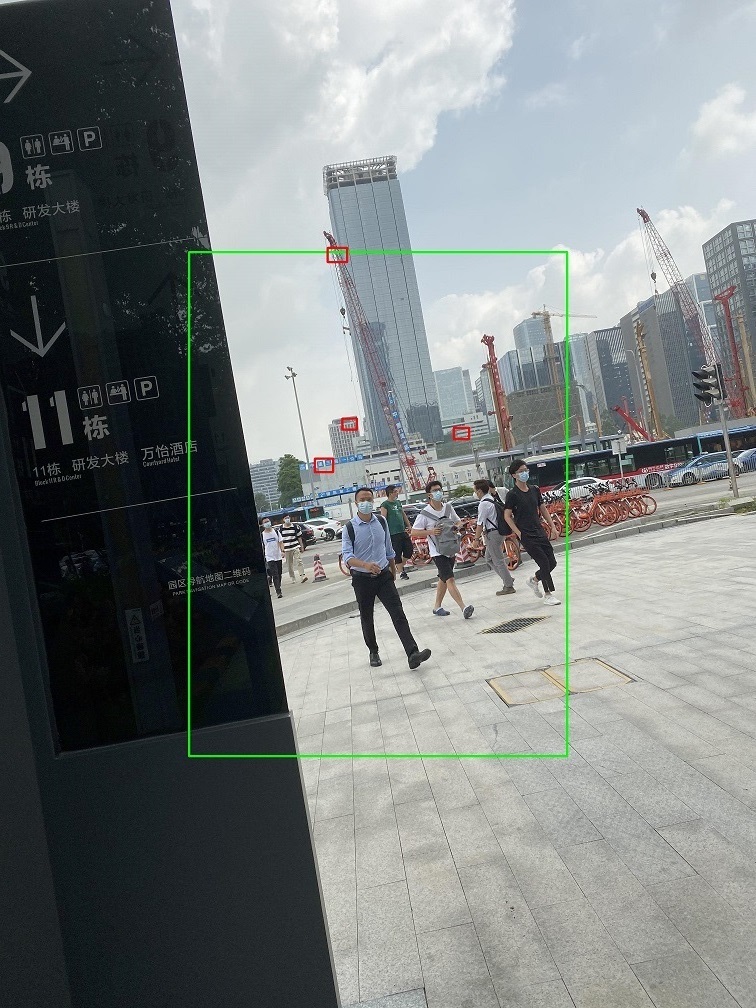}}&
 
\includegraphics[width=0.140\linewidth,height=0.11\linewidth]{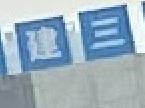}&
\includegraphics[width=0.140\linewidth,height=0.11\linewidth]{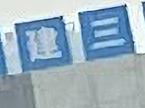}&
\includegraphics[width=0.140\linewidth,height=0.11\linewidth]{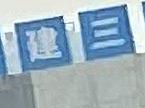}&
\includegraphics[width=0.140\linewidth,height=0.11\linewidth]{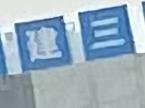}&
\includegraphics[width=0.140\linewidth,height=0.11\linewidth]{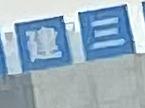}&
\includegraphics[width=0.140\linewidth,height=0.11\linewidth]{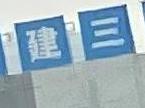}\\
&
\includegraphics[width=0.140\linewidth,height=0.11\linewidth]{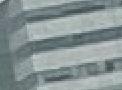}&
\includegraphics[width=0.140\linewidth,height=0.11\linewidth]{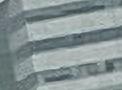}&
\includegraphics[width=0.140\linewidth,height=0.11\linewidth]{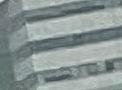}&
\includegraphics[width=0.140\linewidth,height=0.11\linewidth]{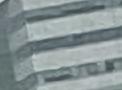}&
\includegraphics[width=0.140\linewidth,height=0.11\linewidth]{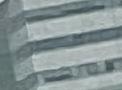}&
\includegraphics[width=0.140\linewidth,height=0.11\linewidth]{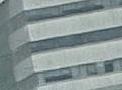}\\

\\
\small Reference  &Bicubic &RCAN  &RSRGAN &CSNLN &TTSR &Ours\\
\multirow{2}[2]{0.17\linewidth}[20.5mm]{ \includegraphics[width=1\linewidth]{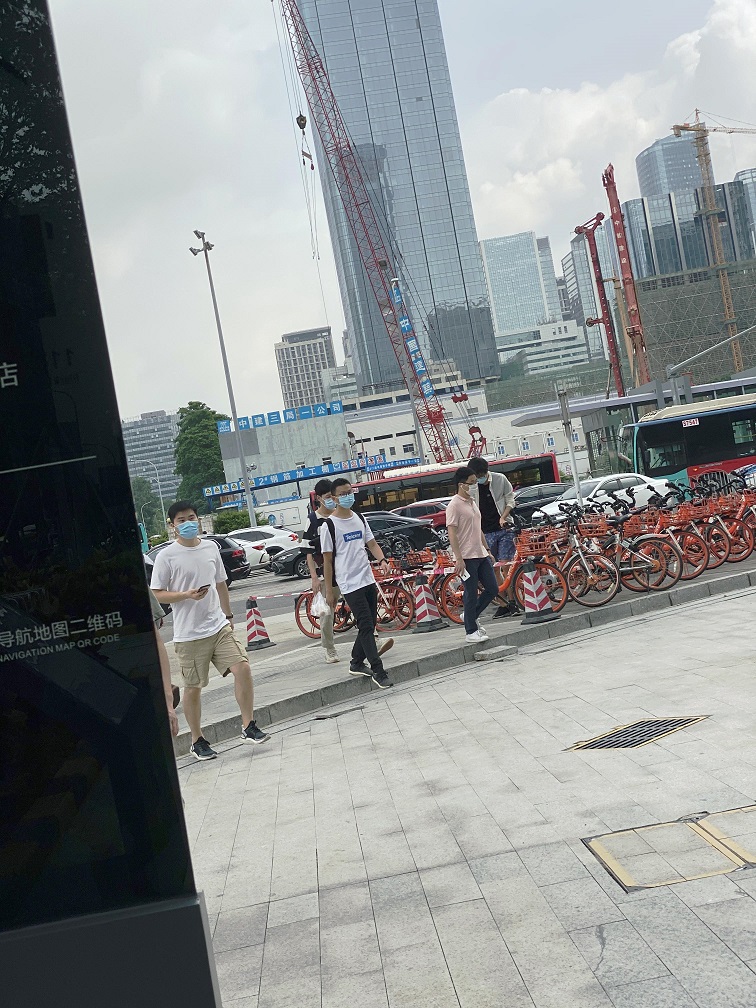}}&
 
\includegraphics[width=0.140\linewidth,height=0.11\linewidth]{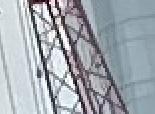}&
\includegraphics[width=0.140\linewidth,height=0.11\linewidth]{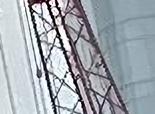}&
\includegraphics[width=0.140\linewidth,height=0.11\linewidth]{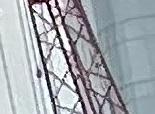}&
\includegraphics[width=0.140\linewidth,height=0.11\linewidth]{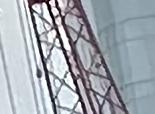}&
\includegraphics[width=0.140\linewidth,height=0.11\linewidth]{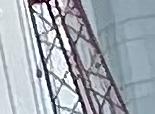}&
\includegraphics[width=0.140\linewidth,height=0.11\linewidth]{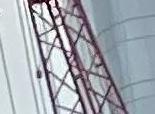}\\
&
\includegraphics[width=0.140\linewidth,height=0.11\linewidth]{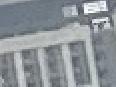}&
\includegraphics[width=0.140\linewidth,height=0.11\linewidth]{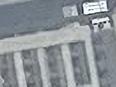}&
\includegraphics[width=0.140\linewidth,height=0.11\linewidth]{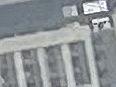}&
\includegraphics[width=0.140\linewidth,height=0.11\linewidth]{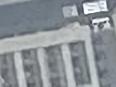}&
\includegraphics[width=0.140\linewidth,height=0.11\linewidth]{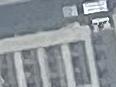}&
\includegraphics[width=0.140\linewidth,height=0.11\linewidth]{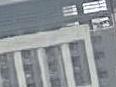}\\
\end{tabular}
\caption{More qualitative comparisons on the CameraFusion dataset. The green box indicates the overlapped FoV area between Input and Ref. Zoom-in for details. }
\label{fig:Our_1}
\end{figure*}

\begin{figure*}[t]
\centering
\small
\begin{tabular}{@{}c@{\hspace{0.4mm}}c@{\hspace{0.4mm}}c@{\hspace{0.4mm}}c@{\hspace{0.4mm}}c@{\hspace{0.4mm}}c@{\hspace{0.4mm}}c@{}}

\small Input  &Bicubic &RCAN &RSRGAN &CSNLN &TTSR &Ours\\
\multirow{2}[2]{0.25\linewidth}[17mm]{ \includegraphics[width=\linewidth]{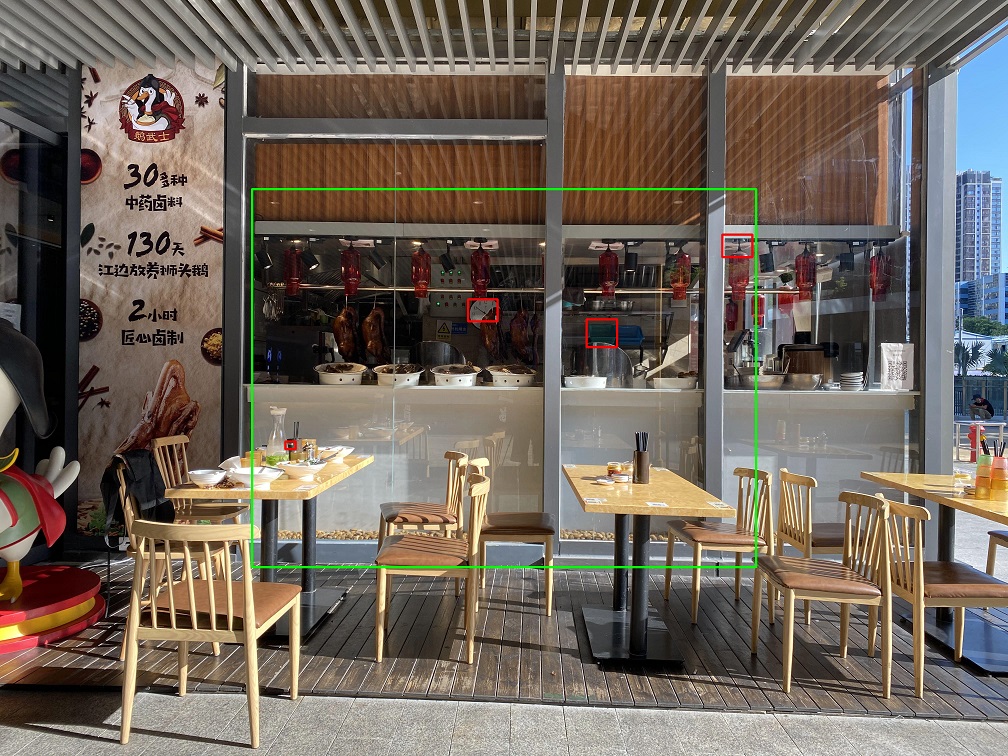}}&
\includegraphics[width=0.12\linewidth,height=0.09\linewidth]{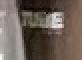}&
\includegraphics[width=0.12\linewidth,height=0.09\linewidth]{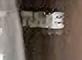}&
\includegraphics[width=0.12\linewidth,height=0.09\linewidth]{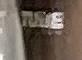}&
\includegraphics[width=0.12\linewidth,height=0.09\linewidth]{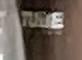}&
\includegraphics[width=0.12\linewidth,height=0.09\linewidth]{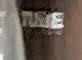}&
\includegraphics[width=0.12\linewidth,height=0.09\linewidth]{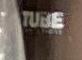}\\
&
\includegraphics[width=0.12\linewidth,height=0.09\linewidth]{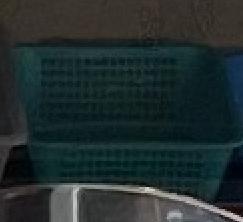}&
\includegraphics[width=0.12\linewidth,height=0.09\linewidth]{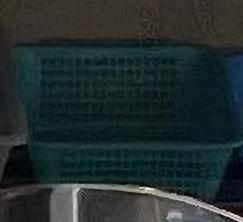}&
\includegraphics[width=0.12\linewidth,height=0.09\linewidth]{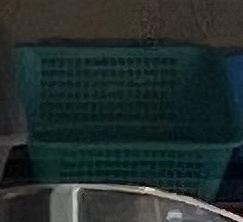}&
\includegraphics[width=0.12\linewidth,height=0.09\linewidth]{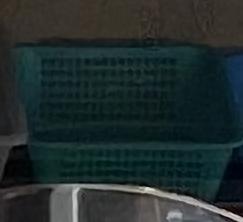}&
\includegraphics[width=0.12\linewidth,height=0.09\linewidth]{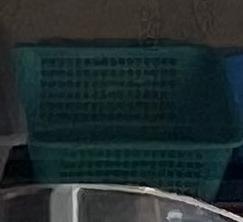}&
\includegraphics[width=0.12\linewidth,height=0.09\linewidth]{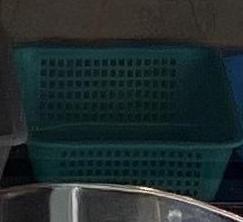}\\

\\
\small Reference  &Bicubic &RCAN &RSRGAN &CSNLN &TTSR &Ours\\
\multirow{2}[2]{0.25\linewidth}[17mm]{ \includegraphics[width=\linewidth]{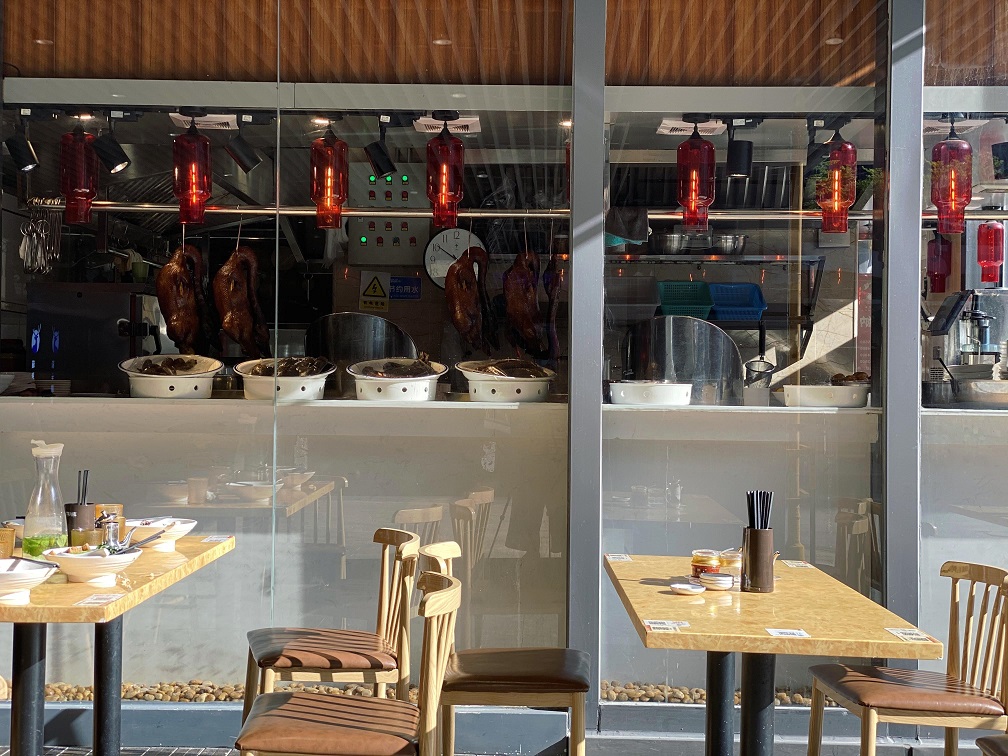}}&
\includegraphics[width=0.12\linewidth,height=0.09\linewidth]{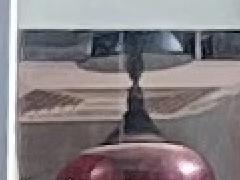}&
\includegraphics[width=0.12\linewidth,height=0.09\linewidth]{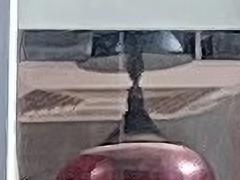}&
\includegraphics[width=0.12\linewidth,height=0.09\linewidth]{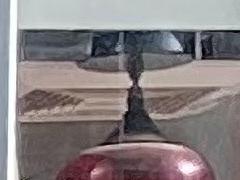}&
\includegraphics[width=0.12\linewidth,height=0.09\linewidth]{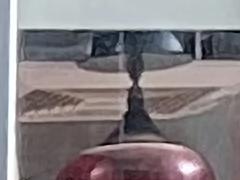}&
\includegraphics[width=0.12\linewidth,height=0.09\linewidth]{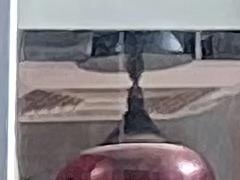}&
\includegraphics[width=0.12\linewidth,height=0.09\linewidth]{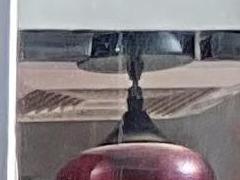}\\
&
\includegraphics[width=0.12\linewidth,height=0.09\linewidth]{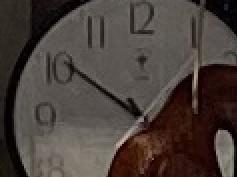}&
\includegraphics[width=0.12\linewidth,height=0.09\linewidth]{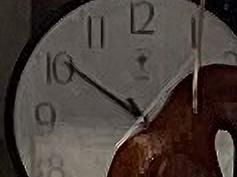}&
\includegraphics[width=0.12\linewidth,height=0.09\linewidth]{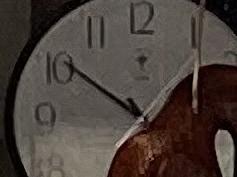}&
\includegraphics[width=0.12\linewidth,height=0.09\linewidth]{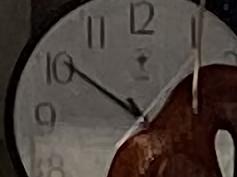}&
\includegraphics[width=0.12\linewidth,height=0.09\linewidth]{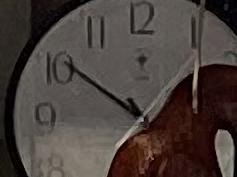}&
\includegraphics[width=0.12\linewidth,height=0.09\linewidth]{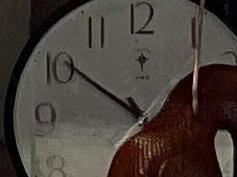}\\
\end{tabular}

\caption{More qualitative comparisons on the CameraFusion dataset. The green box indicates the overlapped FoV area between Input and Ref. Zoom-in for details. }
\label{fig:Our_2}
\end{figure*}
 
\end{document}